\documentclass[runningheads]{llncs}
\usepackage{graphicx}
\graphicspath{ {Images/} }
\usepackage{comment}
\usepackage{amsmath,amssymb} 
\usepackage{color}
\usepackage{multirow}
\usepackage{subcaption}
\usepackage{algorithm}
\usepackage{algorithmic}
\usepackage{lipsum}

\usepackage[width=122mm,left=12mm,paperwidth=146mm,height=193mm,top=12mm,paperheight=217mm]{geometry}

\newcommand{\ie}{\emph{i.e.}~}
\newcommand{\cf}{\emph{cf.}~}
\newcommand{\etal}{\emph{et al.}~}
\newcommand{\vect}[1]{\boldsymbol{#1}}

\begin{document}
\pagestyle{headings}
\mainmatter
\def\ECCVSubNumber{0000}  

\title{Fairness by Learning Orthogonal Disentangled Representations} 


\titlerunning{Fairness by Learning Orthogonal Disentangled Representations}
\author{Mhd Hasan Sarhan\inst{1,2} \and
Nassir Navab\inst{1,3} \and
Abouzar Eslami\inst{1} \and
Shadi Albarqouni\inst{1,4}}
\authorrunning{M.H. Sarhan et al.}
%

\institute{Computer Aided Medical Procedures, Technical University of Munich, Munich, Germany \and
Carl Zeiss Meditec AG, Munich, Germany \and
Computer Aided Medical Procedures, Johns Hopkins University, Baltimore, USA \and
Computer Vision Lab, ETH Zurich, Switzerland
}

\maketitle

\begin{abstract}
Learning discriminative powerful representations is a crucial step for machine learning systems. Introducing invariance against arbitrary nuisance or sensitive attributes while performing well on specific tasks is an important problem in representation learning. This is mostly approached by purging the sensitive information from learned representations. In this paper, we propose a novel disentanglement approach to invariant representation problem. We disentangle the meaningful and sensitive representations by enforcing orthogonality constraints as a proxy for independence. We explicitly enforce the meaningful representation to be agnostic to sensitive information by entropy maximization. The proposed approach is evaluated on five publicly available datasets and compared with state of the art methods for learning fairness and invariance achieving the state of the art performance on three datasets and comparable performance on the rest. Further, we perform an ablative study to evaluate the effect of each component.


\end{abstract}

\section{Introduction}
Learning representations that are useful for downstream tasks yet robust against arbitrary nuisance factors is a challenging problem. Automated systems powered by machine learning techniques are corner stones for decision support systems such as granting loans, advertising, and medical diagnostics. Deep neural networks learn powerful representations that encapsulate the extracted variations in the data. Since these networks learn from historical data, they are prone to represent the past biases and the learnt representations might contain information that were not intended to be released. This has raised various concerns regarding fairness, bias and discrimination in statistical inference algorithms~\cite{mehrabi2019survey}. The European union has recently released their "Ethics guidelines for trustworthy AI" report \footnote{Ethics guidelines for trustworthy AI, \url{https://ec.europa.eu/digital-single-market/en/news/ethics-guidelines-trustworthy-ai}} where it is stated that unfairness and biases must be avoided. \\\\
Since a few years, the community has been investigating to learn a latent representation $\vect{z}$ that well describes a target observed variable $\vect{y}$ (e.g. Annual salary) while being robust against a sensitive attribute $\vect{s}$ (e.g. Gender or race). This nuisance could be independent from the target task which is termed as a domain adaptation problem. One example is the identification of faces $\vect{y}$ regardless of the illumination conditions $\vect{s}$. In the other case termed fair representation learning $\vect{s}$ and $\vect{y}$ are not independent. This could be the case with $\vect{y}$ being the credit risk of a person while $\vect{s}$ is age or gender. Such relation between these variables could be due to past biases that are inherently in the data. This independence is assumed to hold when building fair classification models. Although this assumption is over-optimistic as these factors are probably not independent, we wish to find a representation $\vect{z}$ that is independent from $\vect{s}$ which justifies the usage of such a prior belief~\cite{moyer2018invariant}.
This is mostly approached by approximations of mutual information scores between $\vect{z}$ and $\vect{s}$ and force the two variables to minimize this score either in an adversarial~\cite{xie2017controllable,madras2018learning} or non-adversarial~\cite{louizos2015fairAE,moyer2018invariant} manner. These methods while performing well on various datasets, are still limited by either convergence instability problems in case of adversarial solutions or hindered performance compared to the adversarial counterpart.
Learning disentangled representations has been proven to be beneficial to learning fairer representations compared to general purpose representations~\cite{locatello2019fairness}. 
We use this concept to disentangle the components of the learned representations. Moreover, we treat the $\vect{s}$ and $\vect{y}$ as separate independent generative factors and decompose the learned representation in such a way that each representation holds information related to the respective generative factor. This is achieved by enforcing orthogonality between the representations as a relaxation for the independence constraint. We hypothesize that decomposing the latent code into target code $\vect{z_T}$ and residual sensitive $\vect{z_S}$ code would be beneficial for limiting the leakage of sensitive information into $\vect{z_T}$ by redirecting it to $\vect{z_S}$ while keeping it informative about some target task that we are interested in.

We propose a framework for learning invariant fair representations by decomposing learned representations into target and residual/sensitive representations. We impose disentanglement on the components of each code and impose orthogonality constraint on the two learned representations as a proxy for independence. The learned target representation is explicitly enforced to be agnostic to sensitive information by maximizing the entropy of sensitive information in $\vect{z_T}$. 

Our contributions are three-folds:
\begin{itemize}
    \item Decomposition of target and sensitive data into two orthogonal representations to promote better mitigation of sensitive information leakage.
    \item Promote disentanglement property to split hidden generative factors of each learned code.
    \item Enforce the target representation to be agnostic of sensitive information by maximizing the entropy.
\end{itemize}

\section{Related work}
Learning fair and invariant representations has a long history. Earlier strategies involved changing the examples to ensure fair representation of the all groups. This relies on the assumption that equalized opportunities in the training test would generalize to the test set. Such techniques are referred to as data massaging techniques~\cite{kamiran2009cassifywithoutdiscrimination,pedreshi2008discrimination}. These approaches may suffer of under-utilization of data or complications on the logistics of data collection. Later, Zemel~\etal~\cite{zemel2013LFR} have proposed a semi-supervised fair clustering technique to learn a representation space where data points are clustered such that each cluster contains similar proportions of the protected groups. One drawback is that the clustering constraint limits the power of a distributed representation. To solve this, Louizos~\etal~\cite{louizos2015fairAE} have presented the Variational Fair Autoencoder (VFAE) where a model is trained to learn a representation that is informative enough yet invariant to some nuisance variables. This invariance is approached through Maximum Mean Discrepancy (MMD) penalty. The learned sensitive-information-free representation could be later used for any subsequent processing such as classification of a target task. After the success of Generative Adversarial Networks (GANs)~\cite{goodfellow2014generative}, multiple approaches have leveraged this learning paradigm to produce robust invariant representations~\cite{xie2017controllable,zhang2018unwanted,edwards2015censoring,madras2018learning}. The problem setup in these approaches is a minimax game between an encoder that learns a representation for a target task and an adversary that extracts sensitive information from the learned representation. In this case, the encoder minimizes the negative log-likelihood of the adversary while the adversary is forced to extract sensitive information alternatively. While methods relying on adversarial zeros-sum game of negative log-liklihood minimization and maximization perform well in the literature, they sometimes suffer from convergence problems and require additional regularization terms to stabilize the training. To overcome these problems, Roy~\etal~\cite{roy2019mitigating} posed the problem as an adversarial non-zero sum game where the encoder and discriminator have competing objectives that optimize for different metrics. This is achieved by adding an entropy loss that forces the discriminator to be un-informed about sensitive information. It is worth noting that it is argued by~\cite{moyer2018invariant} that adversarial training for fairness and invariance is unnecessary and sometimes leads to counter productive results. Hence, they have approximated the mutual information between the latent representation and sensitive information using a variational upper bound. Lastly, Creager~\etal~\cite{creager2019flexibly} have proposed a fair representation learning model by disentanglement, their model has the advantage of flexibly changing sensitive information at test time and combine multiple sensitive attributes to achieve subgroup fairness.

\section{Methodology}
\begin{figure}[t]
\centering 
\includegraphics[width=\textwidth]{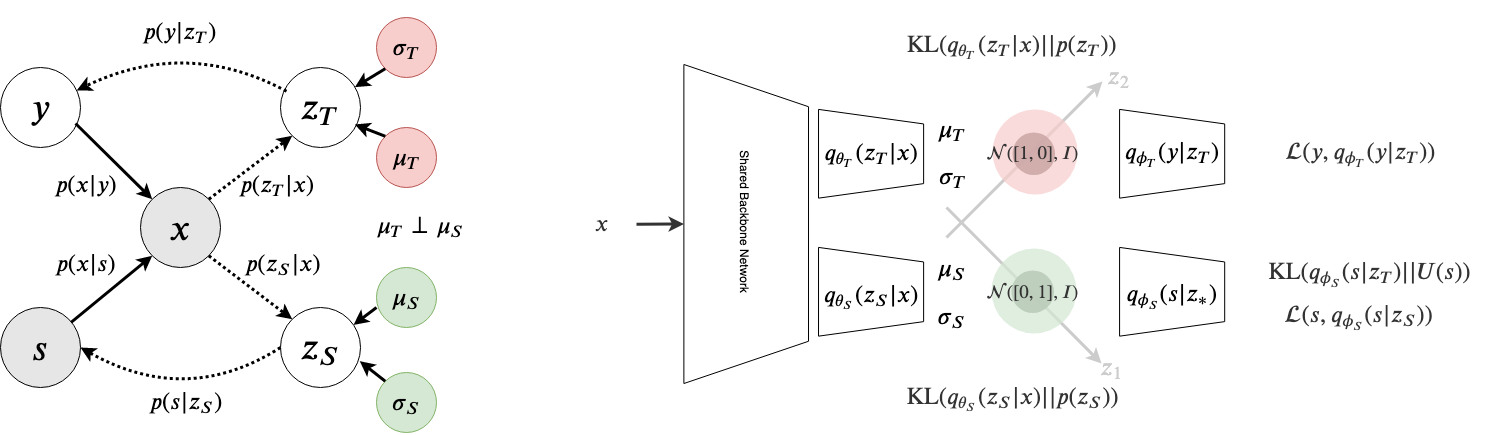}
\caption{Left: The graphical model of our proposed method. Right: Our framework encode the input data to intermediate target and residual (sensitive) representations, parameterized by $\mu$ and $\sigma$. Samples from the estimated posteriors are fed to the discriminators to predict the target and sensitive labels.}
\label{fig:overview}
\end{figure}

let $\mathcal{X}$ be the dataset of individuals from all groups and $\vect{x} \in \mathbb{R}^{D}$ be an input sample.
Each input is associated with a target attribute $\vect{y}=\{y_1,\dots,y_n\} \in \mathbb{R}^{n}$ with $n$ classes, and a sensitive attribute $\vect{s}=\{s_1,\dots,s_m\} \in \mathbb{R}^{m}$ with $m$ classes. Our goal is to learn an encoder that maps input $\vect{x}$ to two low-dimensional representations $\vect{z_T} \in \mathbb{R}^{d_T}$, $\vect{z_S} \in \mathbb{R}^{d_S}$. Ideally $\vect{z_T}$ must contain information regarding target attribute while mitigating leakage about the sensitive attribute and $\vect{z_S}$ contains residual information that is related to the sensitive attribute. 

\subsection{Fairness definition} 
One of the common definition of fairness that has been proposed in the literature~\cite{xie2017controllable,roy2019mitigating,quadrianto2019discovering,barocas-hardt-narayanan} is simply requiring the sensitive information to be statistically independent from the target. Mathematically, the prediction of a classifier $p(\vect{y}|\vect{x})$ must be independent from the sensitive information,\ie $p(\vect{y}|\vect{x}) = p(\vect{y}|\vect{x},\vect{s})$. For example, in the German credit dataset, we need to predict the credit behaviour of the bank account holder regardless the sensitive information, such as gender, age ...etc. In other words, $p(\vect{y}=\text{good credit risk}|\vect{x}, \vect{s}=\text{male})$ should be equal to $p(\vect{y}=\text{good credit risk}|\vect{x}, \vect{s}=\text{female})$. 
The main objective is to learn fair data representations that are i) informative enough for the downstream task, and ii) independent from the sensitive information. 

\subsection{Problem Formulation}

To promote the independence of the generative factors, \ie target and sensitive information, we aim to maximize the log likelihood of the conditional distribution $\log p(\vect{y},\vect{s}|\vect{x})$, where 
\begin{equation}
    p(\vect{y},\vect{s}|\vect{x}) = \frac{p(\vect{y}|\vect{x},\vect{s})p(\vect{x}|\vect{s})p(\vect{s})}{p(\vect{x})} = p(\vect{s}|\vect{x})p(\vect{y}|\vect{x})
\end{equation}

To enforce our aforementioned conditions, we let our model $f(\cdot)$ encode the observed input data $\vect{x}$ into target $\vect{z_T} $ and residual $\vect{z_S}$ representations,
\begin{align}
    p(\vect{y},\vect{s}|\vect{x}) &= p(\vect{y}|\vect{z_T})p(\vect{z_T}|\vect{x})p(\vect{s}|\vect{z_S})p(\vect{z_S}|\vect{x})
\end{align}
and maxmimize the log likelihood given the following constraints; (i) $p(\vect{z_S}|\vect{x})$ is statistically independent from $p(\vect{z_T}|\vect{x})$, and (ii) $\vect{z_T}$ is agnostic to sensitive information $\vect{s}$. Our objective function $J$ can be written as 
\begin{multline}
        J = -\log p(\vect{y},\vect{s}|\vect{x}) \hspace{5pt}
    \text{s.t.} \hspace{5pt} \mathrm{MI}(\vect{z_T}, \vect{z_S}) = 0 \hspace{5pt}\text{and}\hspace{5pt} \mathrm{KL}(p(\vect{s}|\vect{z_T}), \mathcal{U}) = 0
\end{multline}
where $\mathcal{U}(\vect{s})$ is the uniform distribution. 

\subsection{Fairness by Learning Orthogonal and Disentangled Representations} 
As depicted in Fig.~\ref{fig:overview}, our observed data $\vect{x}$ is fed to a shared encoder $f(\vect{x}; \theta)$, then projected into two subspaces producing our target, and residual (sensitive) representations using the encoders; $q_{\theta_T}(\vect{z_T}|\vect{x})$, and $q_{\theta_S}(\vect{z_S}|\vect{x})$, respectively, where $\theta$ is shared parameter, \ie $\theta =\theta_S \cap \theta_T$. Each representation is fed to the corresponding discriminator; target discriminator, $q_{\phi_T}(\vect{y}|\vect{z_T})$, and sensitive discriminator $q_{\phi_S}(\vect{s}|\vect{z_S})$. Both discriminators and encoders are trained in supervised fashion to minimize the following loss, 
\begin{align}
    \mathcal{L}_T(\theta_T, \phi_T) = \mathrm{KL}(p(\vect{y}|\vect{x})~||~q_{\phi_T}(\vect{y}|\vect{z_T})),\\
    \mathcal{L}_S(\theta_S^*, \phi_S) = \mathrm{KL}(p(\vect{s}|\vect{x})~||~q_{\phi_S}(\vect{s}|\vect{z_S})),
\end{align}
where $\theta_S^* = \theta_S \backslash \theta$.
\\
To ensure that our target representation does not encode any leakage of the sensitive information, we follow Roy~\etal\cite{roy2019mitigating} in maximizing the entropy of the sensitive discriminator given the target representation $q_{\phi_S}(\vect{s}|\vect{z_T})$ as
\begin{equation}
     \mathcal{L}_{E}(\phi_S, \theta_T) = \mathrm{KL}(q_{\phi_S}(\vect{s}|\vect{z_T})~||~\mathcal{U}(\vect{s})).
     \label{eq:entropy}
\end{equation}
\\
We relax the independence assumption by enforcing i) disentanglement property, and ii) the orthogonality of the corresponding representations. 

To promote the (i) disentanglement property on the target representation, we first need to estimate the distribution $p(\vect{z_T}|\vect{x})$ and enforce some sort of independence among the latent factors, 
\begin{equation}
    p(\vect{z_T}|\vect{x}) = \frac{p(\vect{x}|\vect{z_T})p(\vect{z_T})}{p(\vect{x})} 
    , \hspace{5pt} \text{s.t.} \hspace{5pt} p(\vect{z_T}) = \prod_{i=1}^{N_T} p(z_T^i).  
\end{equation}

Since $p(z_T|\vect{x})$ is intractable, we employ the Variational Inference, thanks to the re-paramterization trick~\cite{kingma2013auto}, and let our model output the distribution parameters; $\vect{\mu_T}$, and $\vect{\sigma_T}$, and minimize the KL-divergence between posterior $q_{\theta_T}(\vect{z_T}|\vect{x})$ and prior $p(\vect{z_T})$ distributions as 
\begin{equation}
    \mathcal{L}_{\vect{z_T}}(\theta_T) = \mathrm{KL}(q_{\theta_T}(\vect{z_T}|\vect{x}) ~||~ p(\vect{z_T})), 
\end{equation}
where $p(\vect{z_T}) = \prod_{i=1}^{N_T} p(z_T^i) = \mathcal{N}(\textbf{0}, I)$, and $q_{\theta_T}(\vect{z_T}|\vect{x}) = \mathcal{N}(\vect{z_T}; \vect{\mu_T}, diag(\vect{\sigma_T}^2))$. Similarly, we enforce the same constraints on the residual (sensitive) representation $\vect{z_S}$ and minimize the KL-divergence as 
 $\mathcal{L}_{\vect{z_S}}(\theta_S) = \mathrm{KL}(q_{\theta_S}(\vect{z_S}|\vect{x}) ~||~ p(\vect{z_S}))$. 

To enforce the (ii) orthogonality between the target and residual (sensitive) representations,\ie $\vect{\mu_S} \perp \vect{\mu_T}$, we hard code the means of the prior distributions to orthogonal means. In this way, we implicitly enforce the weight parameters to project the representations into orthogonal subspaces. To illustrate this in 2-dimensional space, we set the prior distributions to $p(\vect{z_S}) = \mathcal{N}([0,1]^T, I)$, and $p(\vect{z_T}) = \mathcal{N}([1,0]^T, I)$ (\cf Fig.~\ref{fig:overview}).

To summarize, an additional loss term is introduced to the objective function promoting both Orthogonality and Disentanglement properties, denoted \textit{Orthogonal-Disentangled} loss, 
\begin{equation}
    \mathcal{L}_{OD}(\theta_T, \theta_S) = \mathcal{L}_{\vect{z_T}}(\theta_T) + \mathcal{L}_{\vect{z_S}}(\theta_S).
    \label{eq:od}
\end{equation} 
A variant of this loss without the property of orthogonality, denoted  \textit{Disentangled} loss, is also introduced for the purpose of ablative study (See Sec.~\ref{sec:ablative}). 

\subsection{Overall objective function}
To summarize, our overall objective function is 
\begin{align}
    arg\min_{\theta_T,\theta_S,\phi_T,\phi_S} \mathcal{L}_T(\theta_T, \phi_T) +
    \mathcal{L}_S(\theta_S^*, \phi_S) +
    \lambda_{E}\mathcal{L}_{E}(\phi_S, \theta_T) + \textcolor{blue}{\lambda_{OD}\mathcal{L}_{OD}(\theta_T, \theta_S)}
\end{align}
where $\lambda_E$, and $\lambda_{OD}$ are hyper-parameters to weigh the \textit{Entropy} loss and the \textit{Orthogonal-Disentangled} loss, respectively. A sensitivity analysis on the hyper-parameters is presented in Sec.~\ref{sec:sensitivity}.

\begin{algorithm}
\caption{Learning Orthogonal Disentangled Fair Representations}
\begin{algorithmic}
\REQUIRE Maximum Epochs $E_{max}$, Step size $t_s$, $\lambda_{OD}, \lambda_{E}, \gamma_{OD}, \gamma_{E}, p(\vect{z_T}), p(\vect{z_S})$
\ENSURE $\vect{z_S} \perp \vect{z_T}$ \\
\textbf{Initialize:} $\theta_T, \theta_S, \phi_T, \phi_S \leftarrow \theta_T^{(0)}, \theta_S^{(0)}, \phi_T^{(0)}, \phi_S^{(0)}$\\
\FOR{$t = 1,2, \ldots, E_{max}$}
\STATE $[\vect{\mu_T}, \vect{\sigma_T}] = q_{\theta_T}(\vect{z_T}|\vect{x})$ 
\STATE $[\vect{\mu_S}, \vect{\sigma_S}] = q_{\theta_S}(\vect{z_S}|\vect{x})$
\STATE \textbf{sample} $\vect{z_T} \sim \mathcal{N}(\vect{\mu_T}, diag(\vect{\sigma_T}^2))$ 
\STATE \textbf{sample} $\vect{z_S} \sim \mathcal{N}(\vect{\mu_S}, diag(\vect{\sigma_S}^2))$  
\STATE \textbf{compute} $\mathcal{L}_{\vect{z_T}}(\theta_T) = \mathrm{KL}(q_{\theta_T}(\vect{z_T}|\vect{x}) ~||~ p(\vect{z_T}))$    
\STATE \textbf{compute} $\mathcal{L}_{\vect{z_S}}(\theta_S) = \mathrm{KL}(q_{\theta_S}(\vect{z_S}|\vect{x}) ~||~ p(\vect{z_S}))$
\STATE \textbf{compute} $\mathcal{L}_T(\theta_T, \phi_T) = -\sum p(\vect{y}|\vect{x})\log[q_{\phi_T}(\vect{y}|\vect{z_T})]$
\STATE \textbf{compute} $\mathcal{L}_S(\theta_S^*, \phi_S) = -\sum p(\vect{s}|\vect{x})\log[q_{\phi_S}(\vect{s}|\vect{z_S})]$
\STATE \textbf{compute} $\mathcal{L}_{E}(\phi_S, \theta_T) = \sum q_{\phi_S}(\vect{s}|\vect{z_T})\log[q_{\phi_S}(\vect{s}|\vect{z_T})]$
\STATE \textbf{update} $\lambda_{OD} \leftarrow \lambda_{OD} \gamma_{OD}^{t/t_s}$   
\STATE \textbf{update} $\lambda_{E} \leftarrow \lambda_{E} \gamma_{E}^{t/t_s}$
\STATE $\mathcal{L}_{OD}(\theta_T, \theta_S) \leftarrow \mathcal{L}_{\vect{z_T}}(\theta_T) + \mathcal{L}_{\vect{z_S}}(\theta_S)$  
\STATE $J(\theta_T, \theta_S, \phi_T, \phi_S) = \mathcal{L}_T(\theta_T, \phi_T) +
    \mathcal{L}_S(\theta_S^*, \phi_S) +
    \lambda_{E}\mathcal{L}_{E}(\phi_S, \theta_T) + \textcolor{blue}{\lambda_{OD}\mathcal{L}_{OD}(\theta_T, \theta_S)}$
\STATE \textbf{update} $\theta_T, \theta_S, \phi_T, \phi_S \leftarrow arg\min J(\theta_T, \theta_S, \phi_T, \phi_S)$
\ENDFOR
\RETURN $\theta_T, \theta_S, \phi_T, \phi_S$
\end{algorithmic}
\label{alg:overview}
\end{algorithm}

\section{Experiments}
In this section, the performance of the learned representations using our method will be evaluated and compared against various state of the art methods in the domain.
First, we present the experimental setup by describing the five datasets used for validation, the model implementation details for each dataset, and design of the experiments. We then compare the model performance with state of the art fair representation models on five datasets. We perform an ablative study to monitor the effect of each added component on the overall performance. Lastly, we perform a sensitivity analysis to study the effect of hyper-parameters on the training.

\subsection{Experimental Setup}
\label{sec:experimental-setup}
\paragraph{datasets:}
For evaluating fair classification, we use two datasets from the UCI repository~\cite{dua2017uci}, namely, the German and the Adult datasets. The German credit dataset consists of 1000 samples each with 20 attributes, and the target task is to classify a bank account holder having good or bad credit risk. The sensitive attribute is the gender of the bank account holder. The adult dataset contains 45,222 samples each with 14 attributes. The target task is a binary classification of annual income being more or less than $\$50,000$ and again gender is the sensitive attribute.
\\
To examine the model learned invariance on visual data, we have used the application of illumination invariant face classification. Ideally, we want the representation to contain information about the subject's identity without holding information regarding illumination direction. For this purpose, the extended YaleB dataset is used~\cite{georghiades2001Yaleb}. The dataset contains the face images of 38 subjects under five different light source direction conditions (upper right, lower right, lower left, upper left, and front). The target task is the identification of the subject while the light source condition is considered the sensitive attribute. 
Following Roy~\etal~\cite{roy2019mitigating}, we have created a binary target task from CIFAR-10 dataset~\cite{krizhevsky2009cifar}. The original dataset contains 10 classes we refer to as fine classes, we divide the 10 classes into two categories living and non-living classes and refer to this split as coarse classes. It is expected that living objects have common visual proprieties that differ from non-living ones. The target task is the classification of the coarse classes while not revealing information about the fine classes. With a similar concept, we divide the 100 fine classes of CIFAR-100 dataset into 20 coarse classes that cluster similar concepts into one category. For example, the coarse class 'aquatic mammals' contains the fine classes 'beaver', 'dolphin', 'otter', 'seal', and 'whale'. For the full details of the split, the reader is referred to \cite{roy2019mitigating} or the supplementary materials of this manuscript. The target task for CIFAR-100 is the classification of the coarse classes while mitigating information leakage regarding the sensitive fine classes.

\paragraph{Implementation details:}
For the Adult and German datasets, we follow the setup appeared in~\cite{roy2019mitigating} by having a 1-hidden-layer neural network as encoder, the discriminator has two hidden layer and the target predictor is a logistic regression layer. Each hidden layer contains $64$ units. The size of the representation is $2$. The learning rate for all components is $10^{-3}$ and weight decay is $5\times 10^{-4}$.\\
For the Extended YaleB dataset, we use an experimental setup similar to Xie~\etal~\cite{xie2017controllable} and Louizos~\etal~\cite{louizos2015fairAE} by using the same train/test split strategy. We used $38\times 5=190$ samples for training and $1096$ for testing. The model setup is similar to~\cite{xie2017controllable,roy2019mitigating}, the encoder consisted of one layer, target predictor
is one linear layer and the discriminator is neural network with two hidden layers each contains 100 units. The parameters are trained using Adam optimizer with a learning rate of
$10^{-4}$ and weight decay of $5\times 10^{-2}$.\\
Similar to~\cite{roy2019mitigating}, we employed ResNet-18~\cite{he2016resnet} architecture for training the encoder on the two CIFAR datasets. For the discriminator and target classifiers, we employed a neural network with two hidden layers (256 and 128 neurons). For the encoder, we set the learning rate to $10^{-4}$ and weight decay to $10^{-2}$. For the target and discriminator networks, the learning rate and weight decay were set to $10^{-2}$ and $10^{-3}$,respectively. Adam optimizer~\cite{kingma2014adam} is used in all experiments.

\paragraph{Experiments design:}
We address two questions in the experiments. First, is \textit{how much information about the sensitive attributes is retained in the learned representation $\vect{z_T}$?}. Ideally, $\vect{z_T}$ would not contain any sensitive attribute information. This is evaluated by training a classifier with the same architecture as the discriminator network on sensitive attributes classification task. The closer the accuracy to a naive majority label predictor, the better the model is. This classifier is trained with $\vect{z_T}$ as input after the encoder, target, and discriminator had been trained and freezed. Second, is \textit{how well the learned representation $\vect{z_T}$ performs in identifying target attributes?}. To this end, we train a classifier similar to the target on the learned representation $\vect{z_T}$ to detect the target attributes. We also visualize the representations $\vect{z_T}$ and $\vect{z_S}$ by using their t-SNE projections to show how the learned representations describe target attributes while being agnostic to the sensitive information.

\subsection{Comparison with state of the art}

\begin{table}[t]
\centering
\caption{Results on CIFAR-10 and CIFAR-100 datasets.}
\resizebox{\textwidth}{!}{%
\begin{tabular}{l|c|c|c|c}
\hline
\multirow{2}{*}{}          & \multicolumn{2}{c|}{CIFAR-10}        & \multicolumn{2}{c}{CIFAR-100}       \\ \cline{2-5} 
                           & Target Acc. $\uparrow$ & Sensitive Acc. $\downarrow$ & Target Acc. $\uparrow$ & Sensitive Acc. $\downarrow$\\ \hline
Baseline                   & 0.9775          & 0.2344             & \textbf{0.7199} & 0.3069             \\ 
\hline
Xie et al.~\cite{xie2017controllable} (trade-off \#1) & 0.9752          & 0.2083             & 0.7132          & 0.1543             \\ 
Roy et al.~\cite{roy2019mitigating} (trade-off \#1) & \textbf{0.9778} & 0.2344             & 0.7117          & 0.1688             \\ 
Xie et al.~\cite{xie2017controllable} (trade-off \#2) & 0.9735          & 0.2064             & 0.7040           & 0.1484             \\ 
Roy et al.~\cite{roy2019mitigating} (trade-off \#2) & 0.9679          & 0.2114             & 0.7050           & 0.1643             \\ 
\hline
Ours                       & 0.9725          & \textbf{0.1907}    & 0.7074          & \textbf{0.1447}    \\ \hline
\end{tabular}%
}
\label{tbl:compare}

\end{table}

\begin{figure}[t]
    \centering
    \begin{subfigure}{1\textwidth}
        \includegraphics[width=0.32\textwidth]{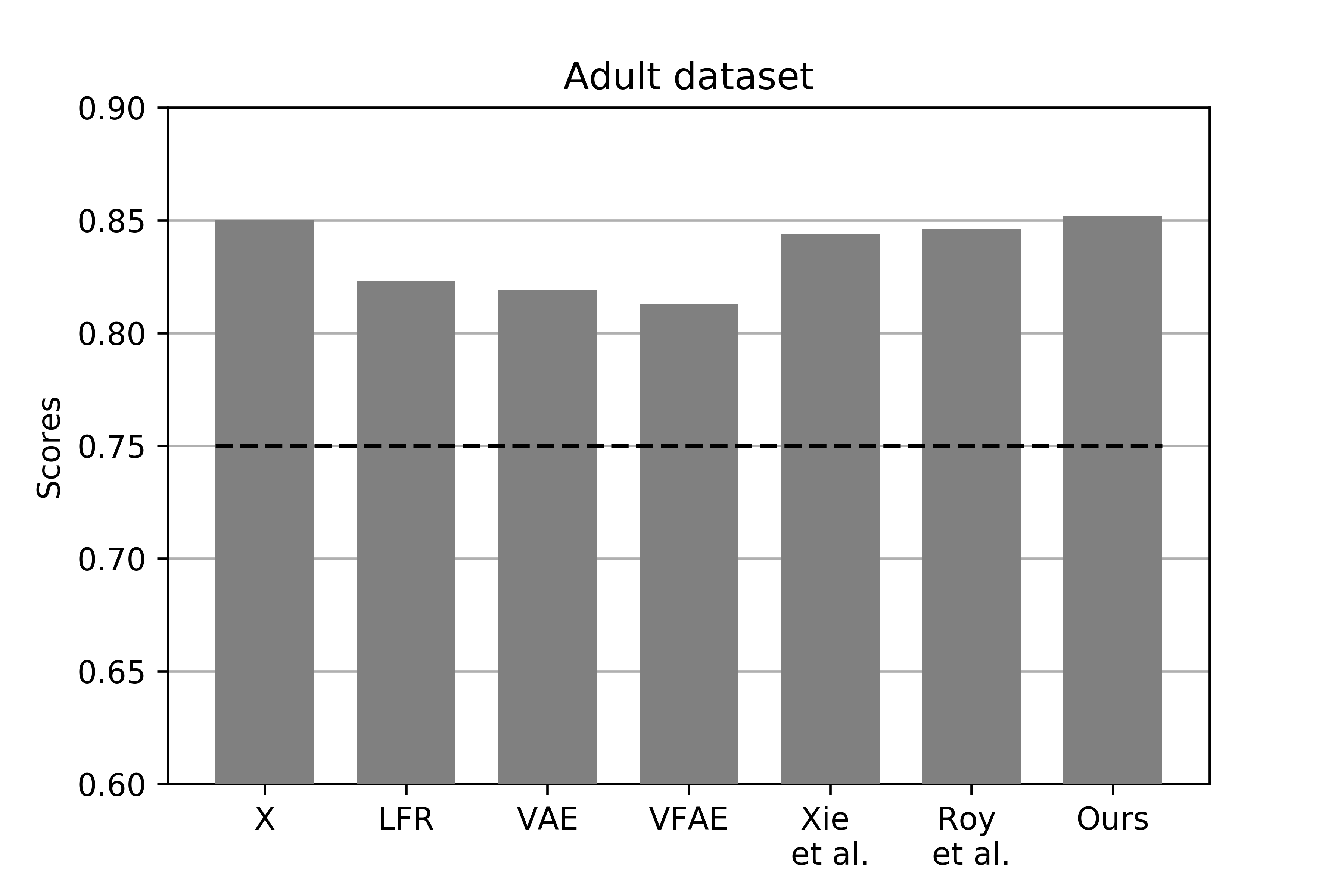}
        \includegraphics[width=0.32\textwidth]{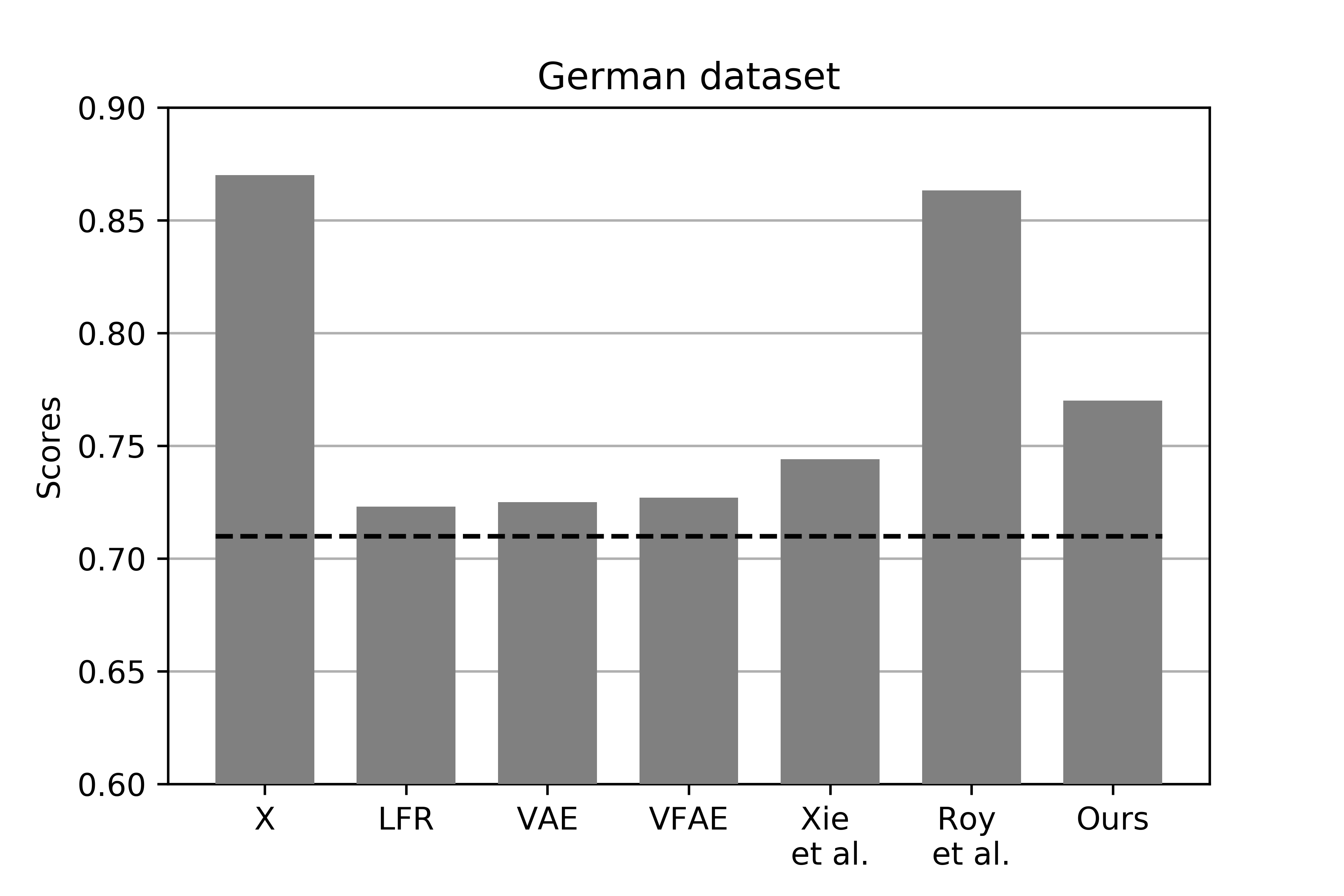}
        \includegraphics[width=0.32\textwidth]{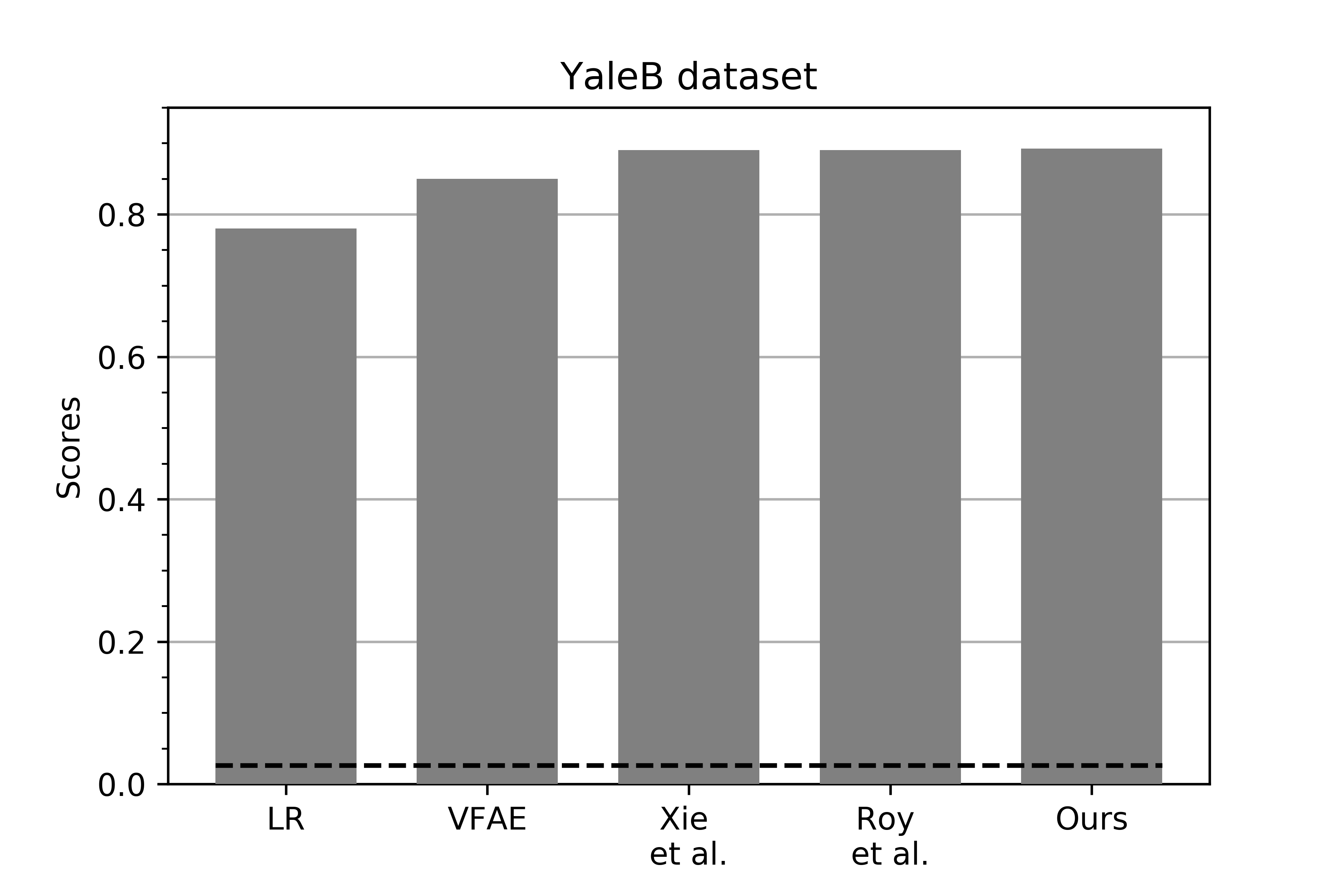}
        \caption{Target attribute classification accuracy.} \label{fig:compare1a}
    \end{subfigure}
    
    \vspace{0.2cm}
    \begin{subfigure}{1\textwidth}
        \includegraphics[width=0.32\textwidth]{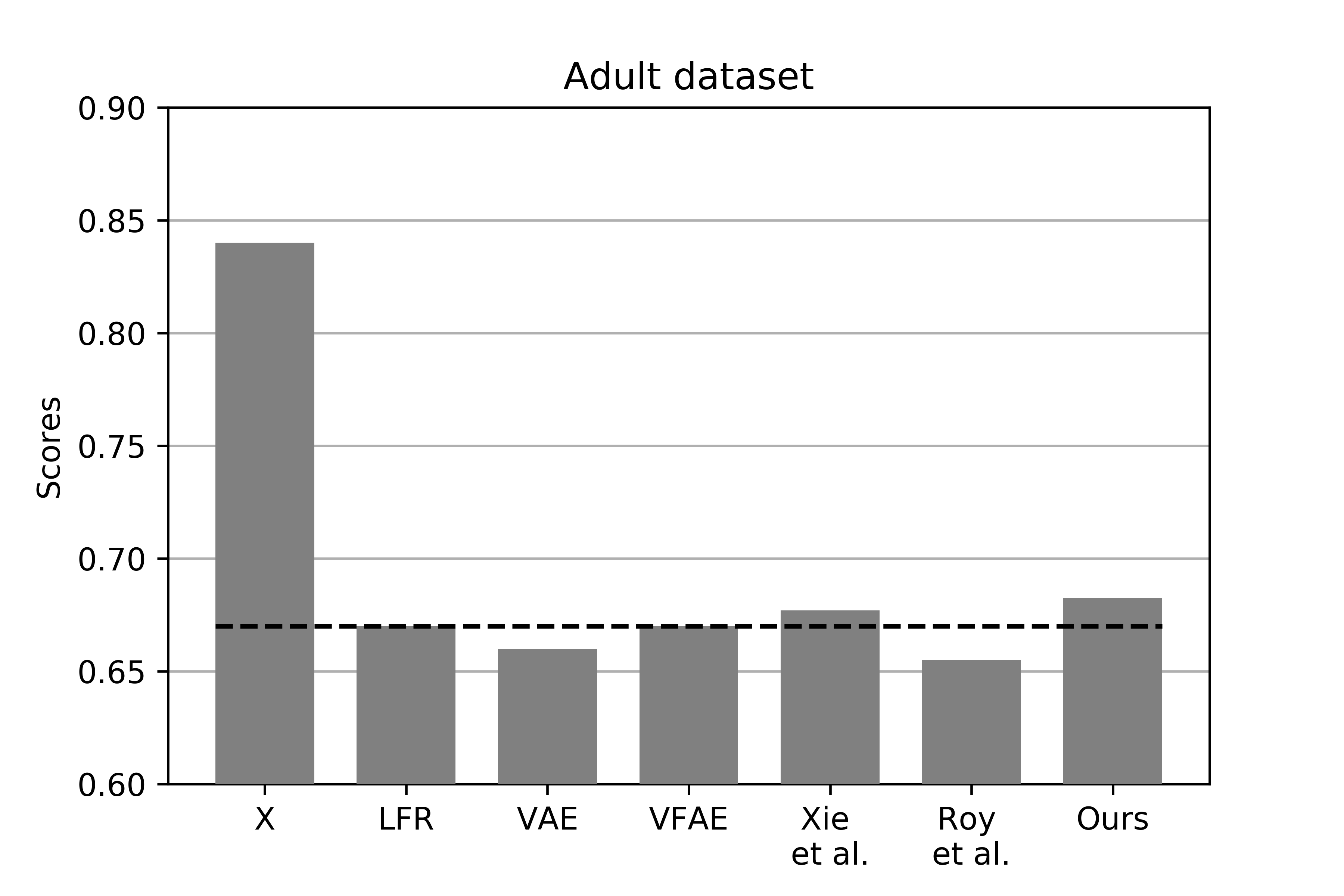}
        \includegraphics[width=0.32\textwidth]{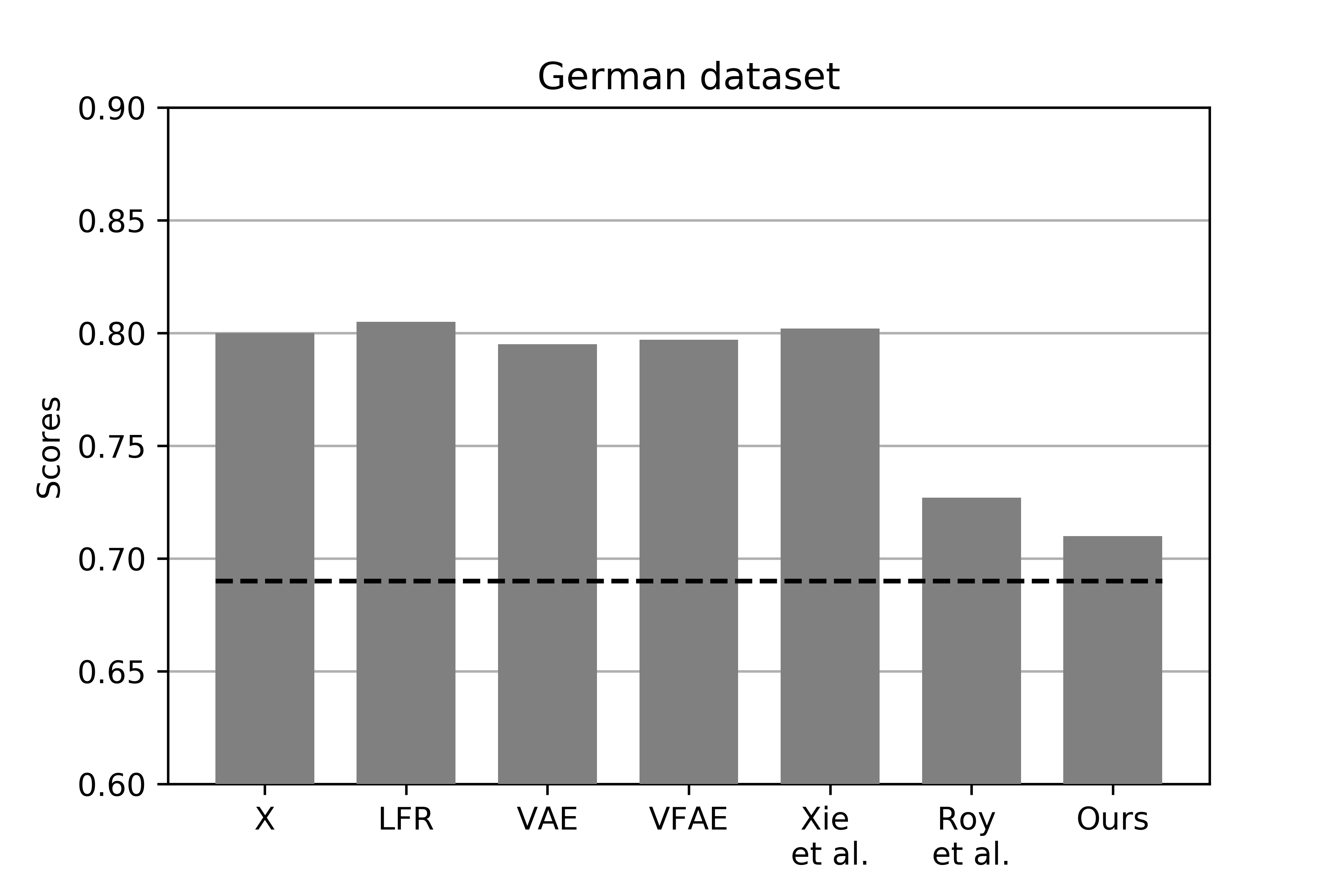}
        \includegraphics[width=0.32\textwidth]{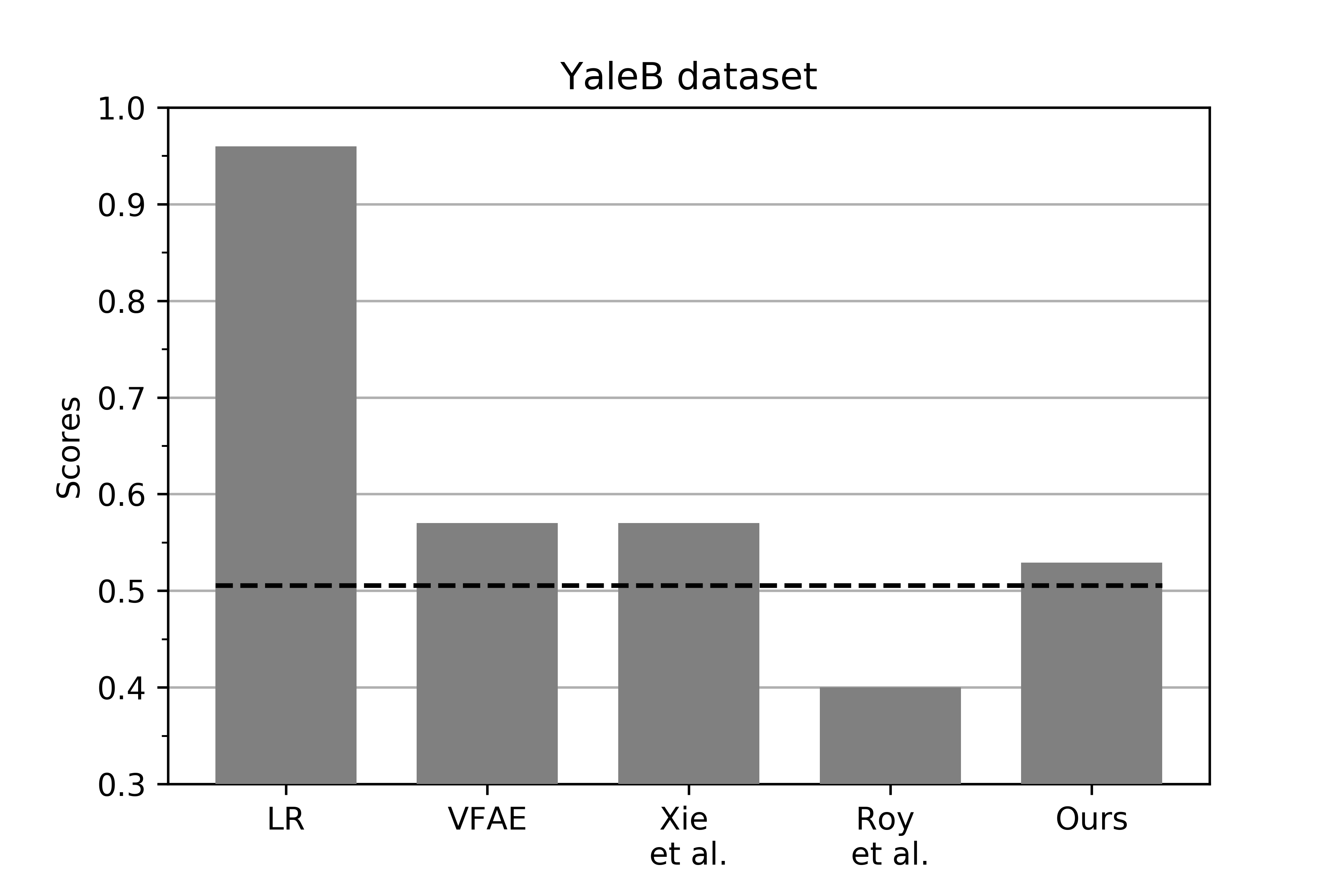}
        \caption{Sensitive attribute classification accuracy.} \label{fig:compare1b}
    \end{subfigure}
    \caption{Results on Adult, German, and extended YaleB datasets. The dashed black line represent a naive majority classifier that predicts the majority label.}
    \label{fig:compare1}
\end{figure}

We compare the proposed approach against various state of the art methods on the five presented datasets. We first train the model with Algorithm~\ref{alg:overview} while changing hyper-parameters between runs.
We choose the best performing model in terms of the trade-off between target and sensitive classification accuracy based on $\vect{z_T}$. We then compare it with various state of the art methods for sensitive information leakage and retaining target information.\\
\paragraph{CIFAR datasets:} We compare the proposed approach with two other state of the art methods on the CIFAR-10 and CIFAR-100 datasets, namely Xie~\etal~\cite{xie2017controllable} and Roy~\etal~\cite{roy2019mitigating}. We examine two different trade-off points of both approaches. The first trade-off point is the one with best target accuracy reported by the model while the second trade-off point is the one with the target accuracy closest to ours for a more fair comparison. The lower the target accuracy in the trade-off the better (lower) the sensitive accuracy is. We can see when the target accuracies are comparable, our model performs better in preventing sensitive information leakage to the representation $\vect{z_T}$. Hence, the proposed method has a better trade-off on the target and sensitive accuracy for both CIFAR-10 and CIFAR-100 datasets. However, the peak target performance is comparable but lower than the peak target performance of the studied methods.
\paragraph{Extended YaleB dataset:} For the illumination invariant classification task on the extended YaleB dataset, the proposed method is compared with the logistic regression baseline (LR), variational fair autoencoder VFAE~\cite{louizos2015fairAE},  Xie~\etal~\cite{xie2017controllable} and Roy~\etal~\cite{roy2019mitigating}. The results are shown in Fig.~\ref{fig:compare1} on the right hand side. The proposed model performs best on the target attribute classification while having the closest performance to the majority classification line (dashed line in Fig.~\ref{fig:compare1}). The majority line is the trivial baseline of predicting the majority label. The closer the sensitive accuracy to the majority line the better the model is in hiding sensitive information from $\vect{z_T}$. This means the learned representation is powerful at identifying subject in the images regardless of illumination conditions. To assess this visually, refer to sec.~\ref{sec:qualitative} for qualitative analysis.

\paragraph{Tabular datasets:} On the Adult and German datasets, we compare with LFR~\cite{zemel2013LFR}, vanilla VAE~\cite{kingma2013auto}, variational fair autoencoder~\cite{louizos2015fairAE}, Xie~\etal~\cite{xie2017controllable} and Roy~\etal~\cite{roy2019mitigating}. The results of these comparisons are shown in Fig.~\ref{fig:compare1}. On the German dataset, we observe a very good performance in hiding sensitive information with $71\%$ accuracy compared to $72.7\%$ in~\cite{roy2019mitigating}. On the target task, the model performs well compared to other models except for \cite{roy2019mitigating} which does marginally better than the rest. On the Adult dataset, our proposed model performs better than the aforementioned models on the target task while leaking slightly more information compared to other methods and the majority line at $67\%$. Our method has $68.26\%$ sensitive accuracy while LFR, VAE, vFAE, Xie~\etal, and Roy~\etal have $67\%$, $66\%$, $67\%$, $67.7\%$, and $65.5\%$ sensitive accuracy, respectively. 

Generally, we observe that the proposed model performs well on all datasets with state of the art performance on visual datasets (CIFAR-10, CIFAR-100, YaleB). This suggests that such a model could lead to more fair/invariant representation without large sacrifices on downstream tasks.

\subsection{Ablative study}
\label{sec:ablative}
\begin{figure}[t]
    \centering
    \includegraphics[width=0.32\textwidth]{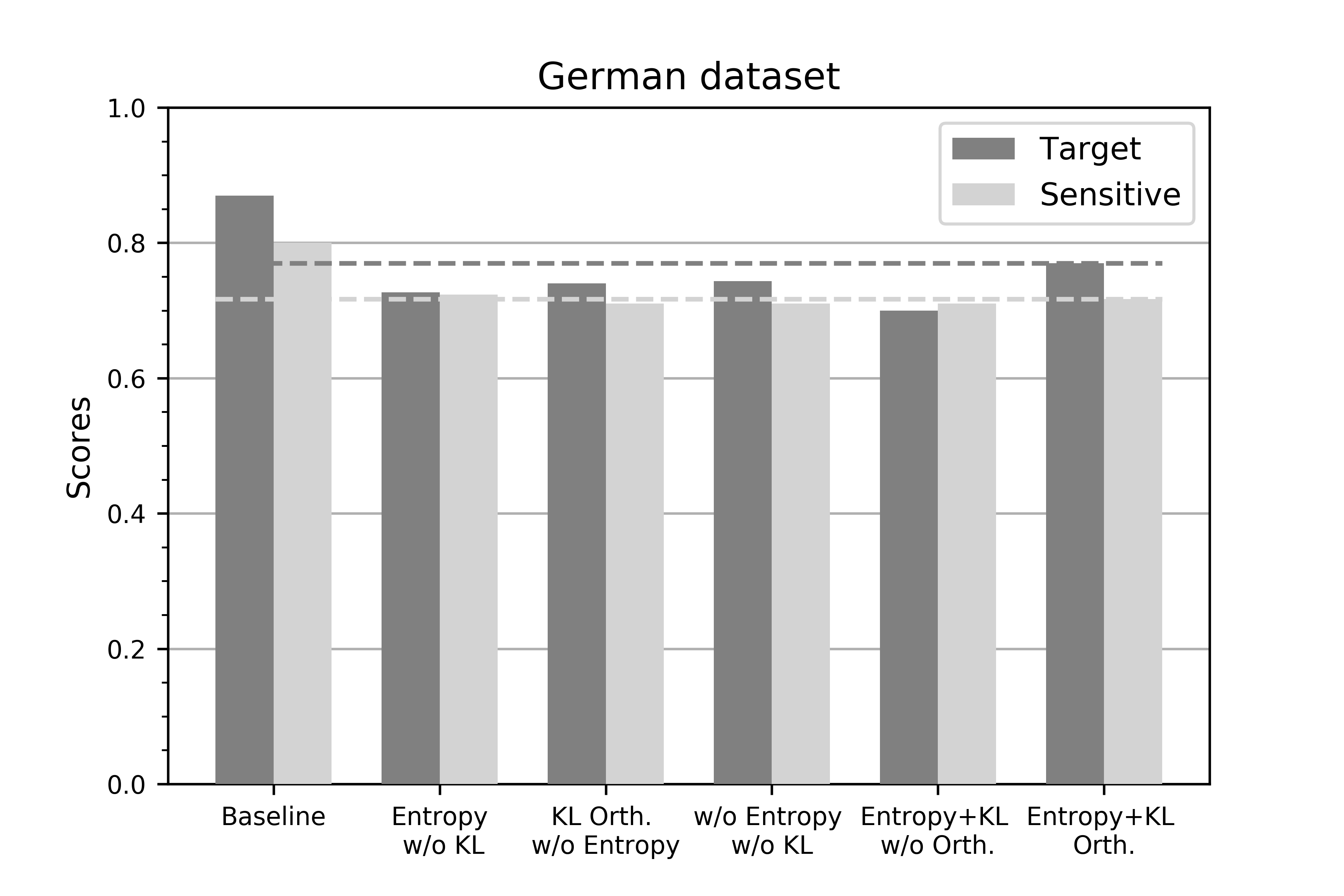}
    \includegraphics[width=0.32\textwidth]{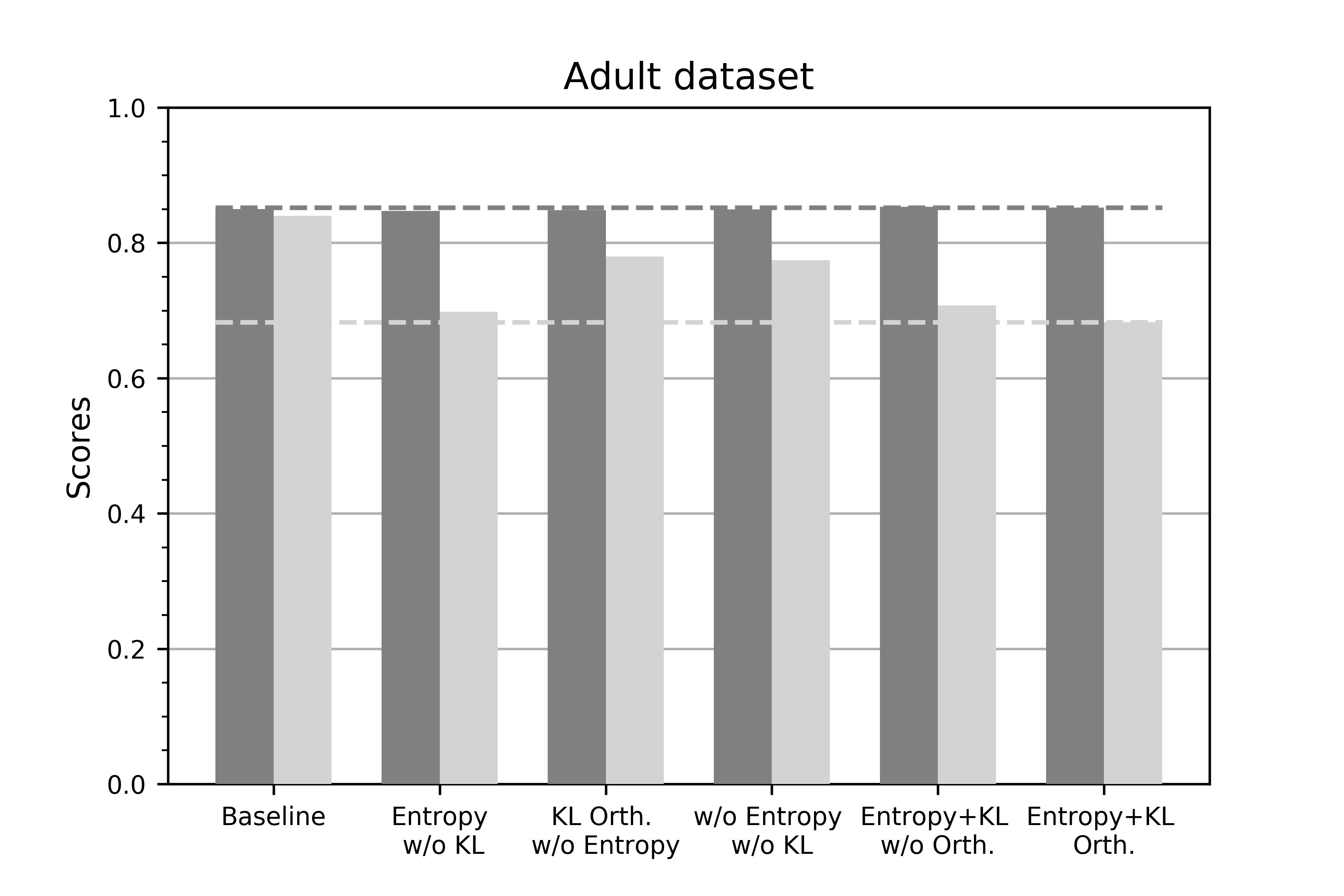}
    \includegraphics[width=0.32\textwidth]{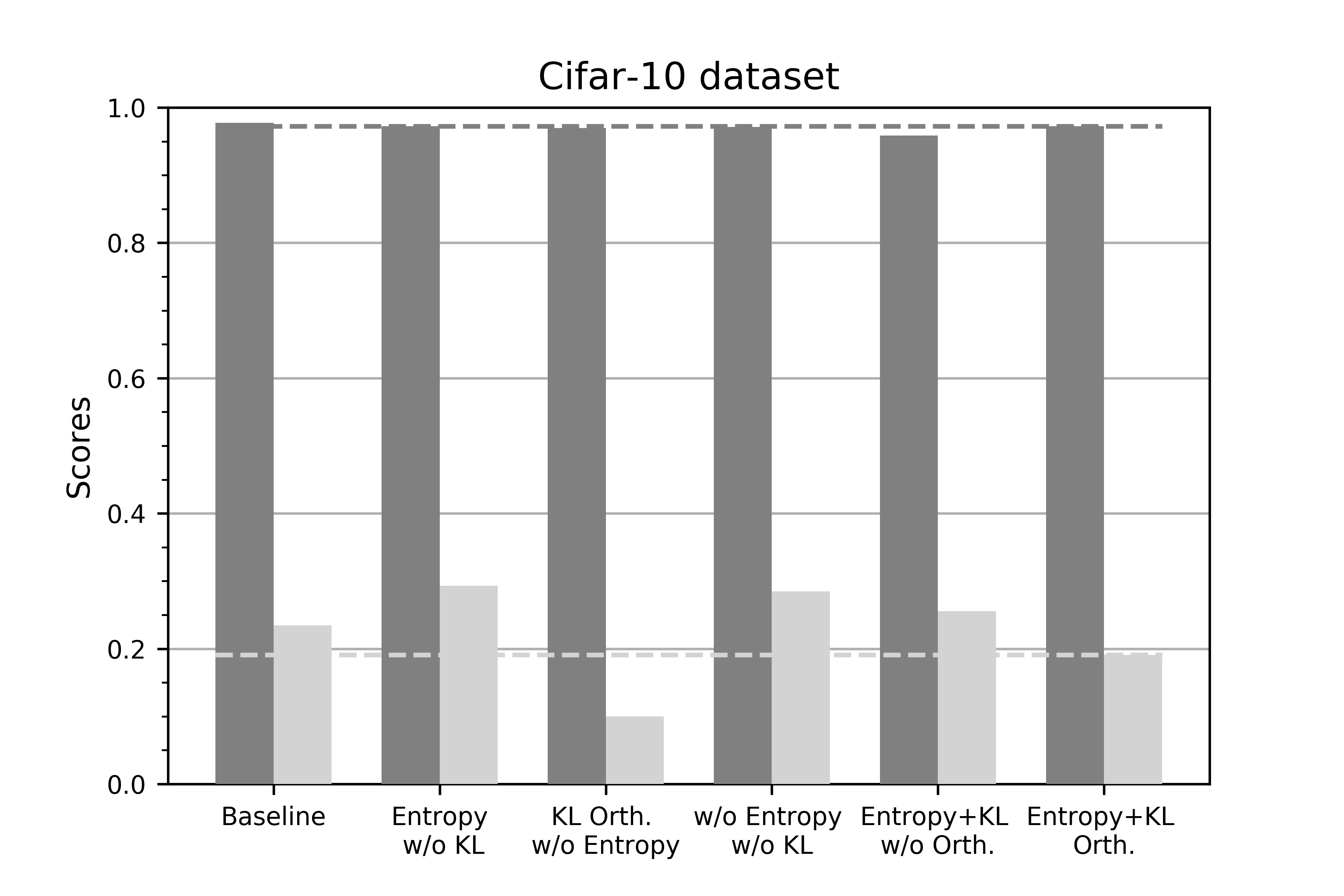}
    \vspace{0.2cm}
    
    \includegraphics[width=0.32\textwidth]{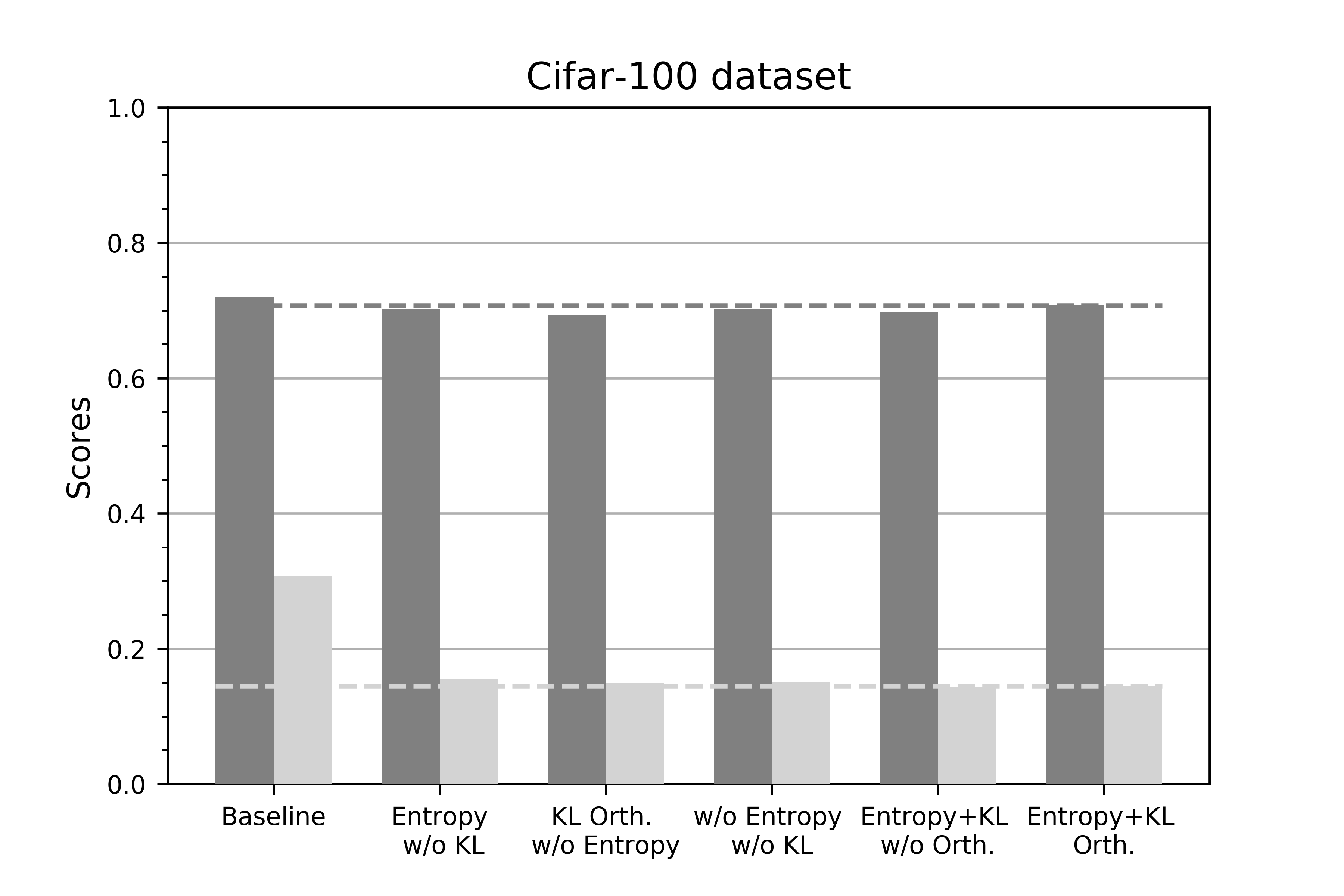}
    \includegraphics[width=0.32\textwidth]{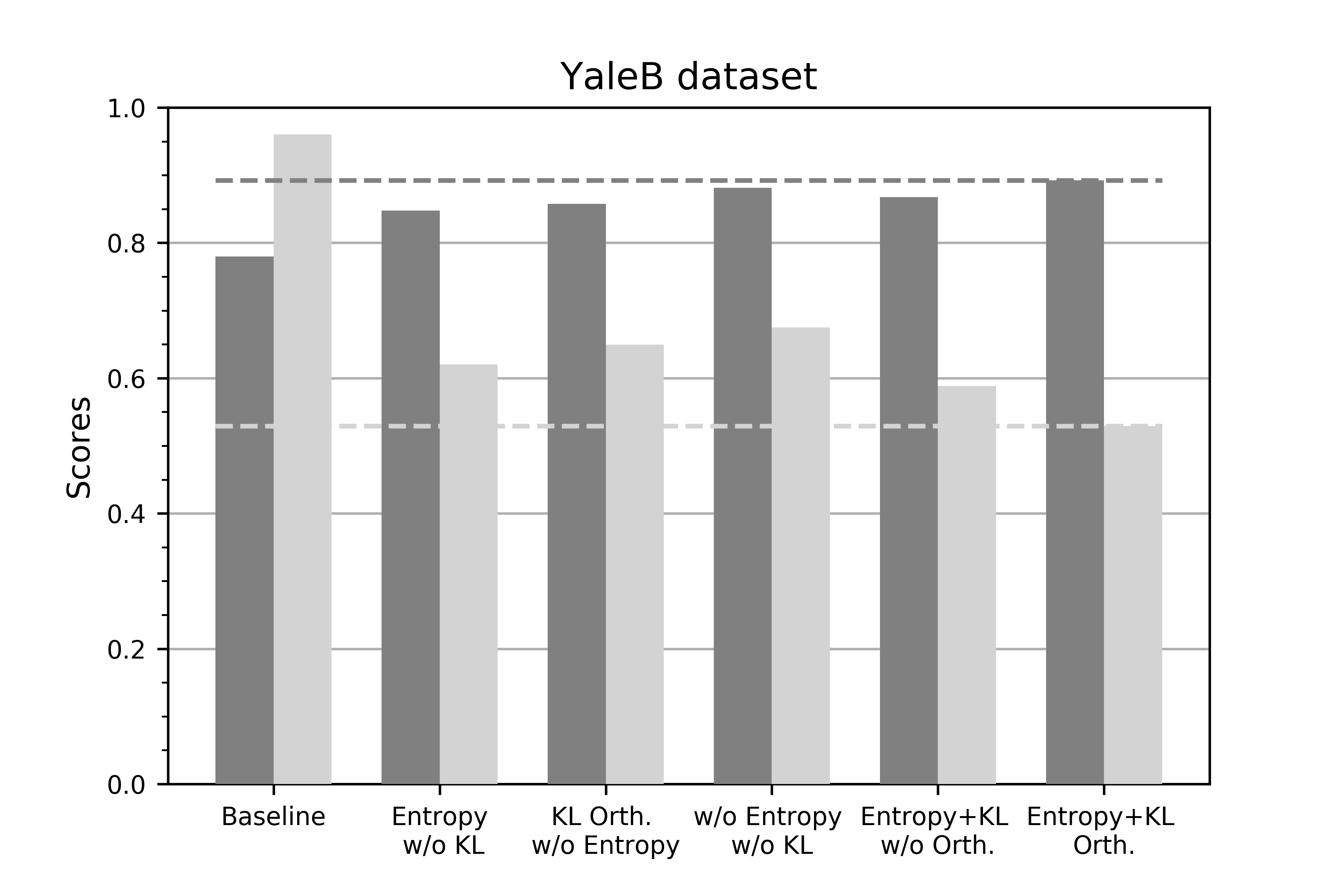}
    \caption{Ablative study. Dark gray and light gray dashed lines represent the accuracy results on the target and sensitive task respectively for the "Entropy + KL Orth." model. }
    \label{fig:ablative}
\end{figure}

In this section, we evaluate the contributions provided in the paper by eliminating parts of the loss function and study how each part affects the training in terms of target and sensitive accuracy. To this end, we used the best performing models after hyper-parameter search when training for all contributions for each dataset. The models are trained with the same settings and architectures described in Sec. \ref{sec:experimental-setup}. We compare five different variations for each model alongside the baseline classifier:
\begin{enumerate}
    \item \textbf{Baseline}: Training a deterministic classifier for the target task and evaluate the information leakage about the sensitive attribute.
    \item \textbf{Entropy w/o KL}: Entropy loss $\mathcal{L}_{E}$ is incorporated (Equation~\ref{eq:entropy}) in the loss while $\mathcal{L}_{OD}$ is not included (Equation~\ref{eq:od}).
    \item \textbf{KL Orth. w/o Entropy}: Entropy loss $\mathcal{L}_{E}$ is not used (Equation~\ref{eq:entropy}) while $\mathcal{L}_{OD}$ is used for target and sensitive representations with orthogonal means (Equation~\ref{eq:od}).
    \item \textbf{w/o Entropy w/o KL}: Neither entropy loss nor KL divergence are used in the loss. This case is similar ti multi-task learning with the tasks being the classification of target and sensitive attributes.
    \item \textbf{Entropy + KL w/o Orth.}: Entropy loss $\mathcal{L}_{E}$ is used and \textit{disentangled loss} is used with similar means. Hence, there might be some disentanglement of generative factors in the components of each latent code but no constraints are applied to force disentanglement of the two representations.
    \item \textbf{Entropy + KL Orth.}: All contributions are included.
\end{enumerate}
The results of the ablative study are shown in Figure~\ref{fig:ablative}. 
\begin{itemize}
    \item For the \textit{sensitive class accuracy}, it is desirable to have a lower accuracy in distinguishing sensitive attributes. Compared to the baseline, we observe that adding entropy loss and orthogonality constraints on the representations lowers the discriminative power of the learned representation regarding sensitive information. This is valid on all studied datasets except for CIFAR-10 where orthogonality constraint without entropy produced better representations for hiding sensitive information with a small drop ($0.26\%$) on the target task performance. In the rest of the cases, having either entropy loss or KL loss only does not bring noticeable performance gains compared to a multi-task learning paradigm. This could be attributed to the fact that orthogonality on its own does not enforce independence of random variables and another constraint is needed to encourage independent latent variables (\ie entropy loss).
    \item Comparing baseline with \textbf{w/o Entropy w/o KL} case answers the important question \textit{"Does multi-task learning with no constraints on representations bring any added value in mitigating sensitive information leakage?"}. In three out of the five studied datasets, it is the case. We can see lower accuracy in identifying sensitive information by using the learned target representation as input to a classifier while having no constraints on the relationship between the sensitive and target representations during the training process of the encoder. Simply, adding an auxiliary classifier to the target classifier and force it to learn information about sensitive attributes hides some sensitive data from the target classifier. 
    \item Regarding \textbf{target accuracy}, the proposed model does not suffer from large drops in target performance when disentangling target from sensitive information. This could be seen by comparing target accuracy between the baseline and \textbf{Entropy+KL Orth.} columns. The largest drop in target performance compared to no privacy baseline is seen on the German dataset. This could be because of the very high dependence between gender and granting good or bad credit to a subject in the dataset and the small amount of subjects in the dataset.

\end{itemize}

\subsection{Qualitative analysis}
\label{sec:qualitative}
\begin{figure}[t]
    \centering
    \begin{subfigure}{0.31\textwidth}
        \includegraphics[width=1\textwidth]{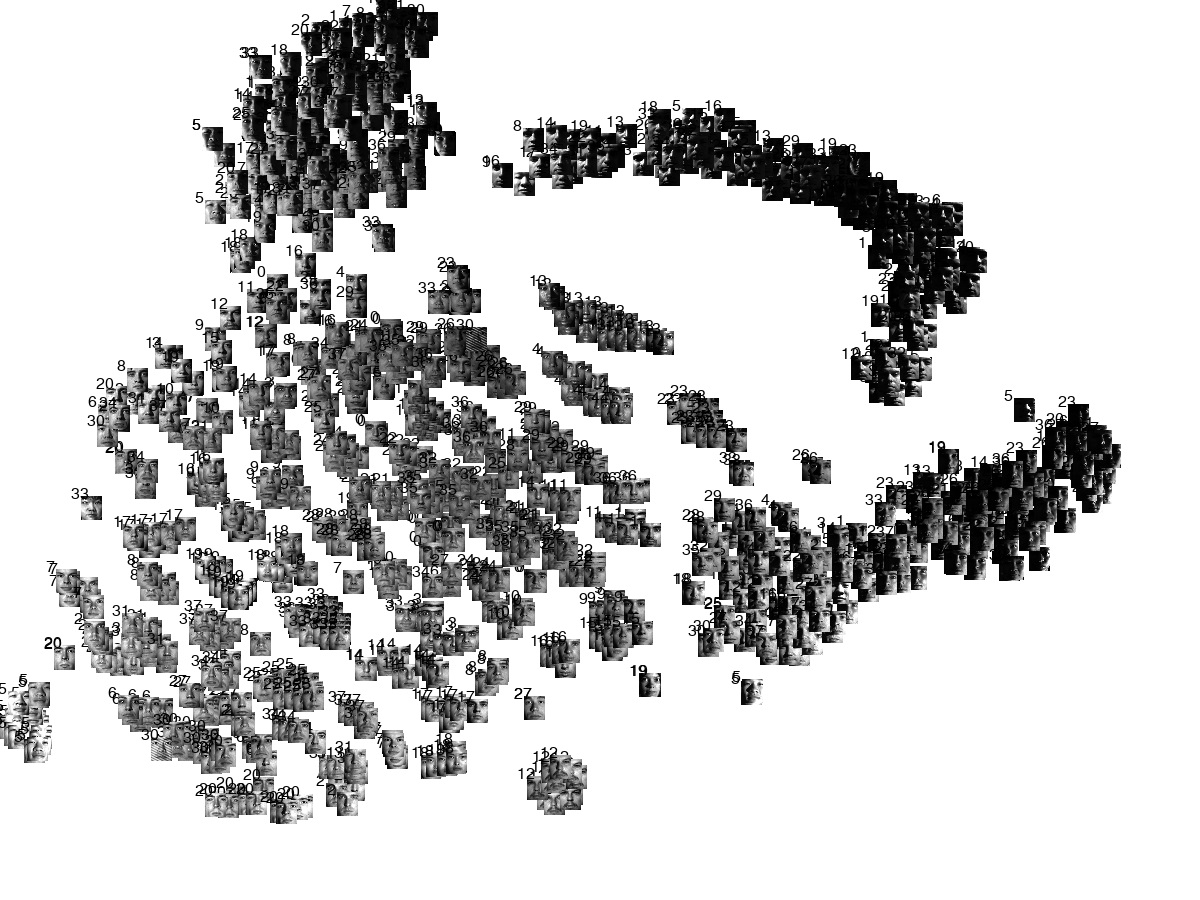}
        \caption{t-SNE on $x$} \label{fig:compare1a}
    \end{subfigure}
    \hfill
    \begin{subfigure}{0.31\textwidth}
        \includegraphics[width=1\textwidth]{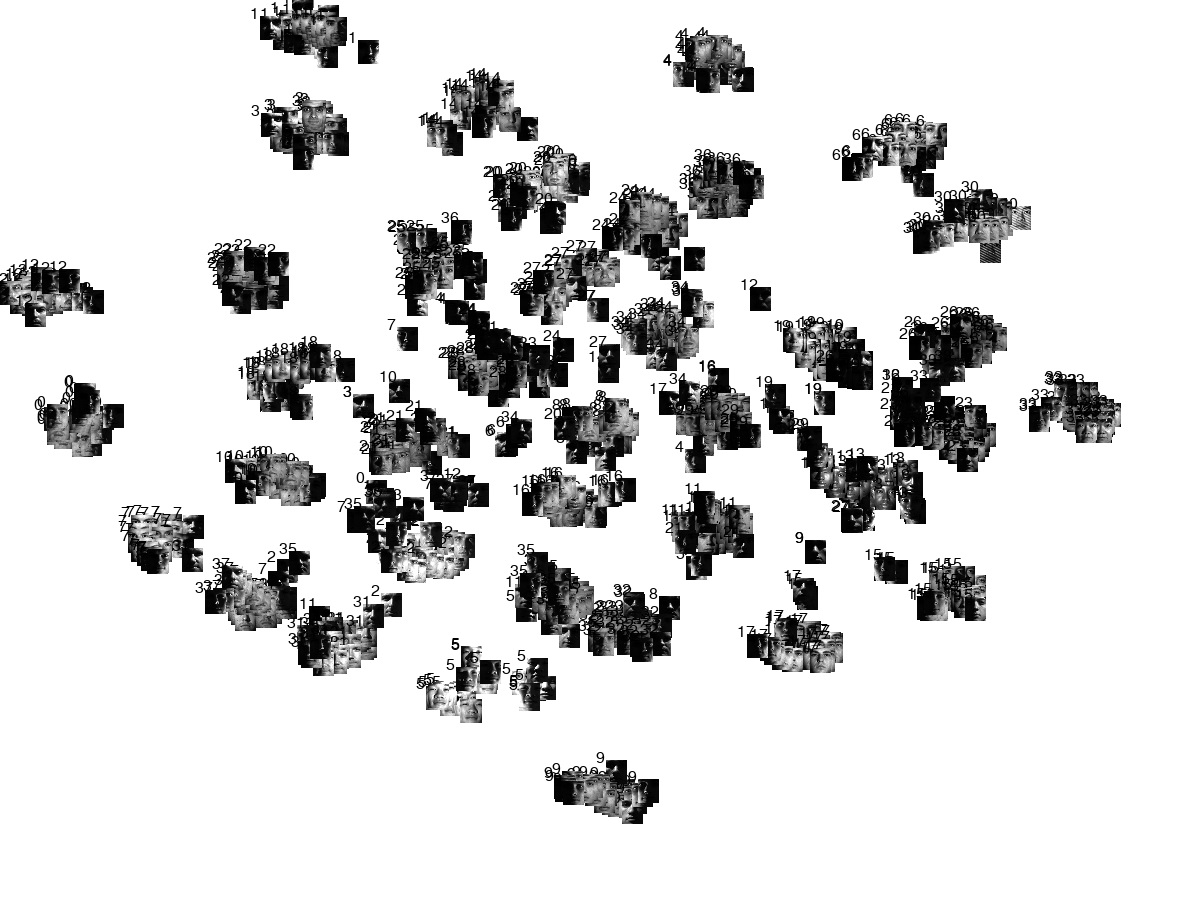}
        \caption{t-SNE on $\vect{z_T}$} \label{fig:compare1a}
    \end{subfigure}
    \hfill
    \begin{subfigure}{0.31\textwidth}
        \includegraphics[width=1\textwidth]{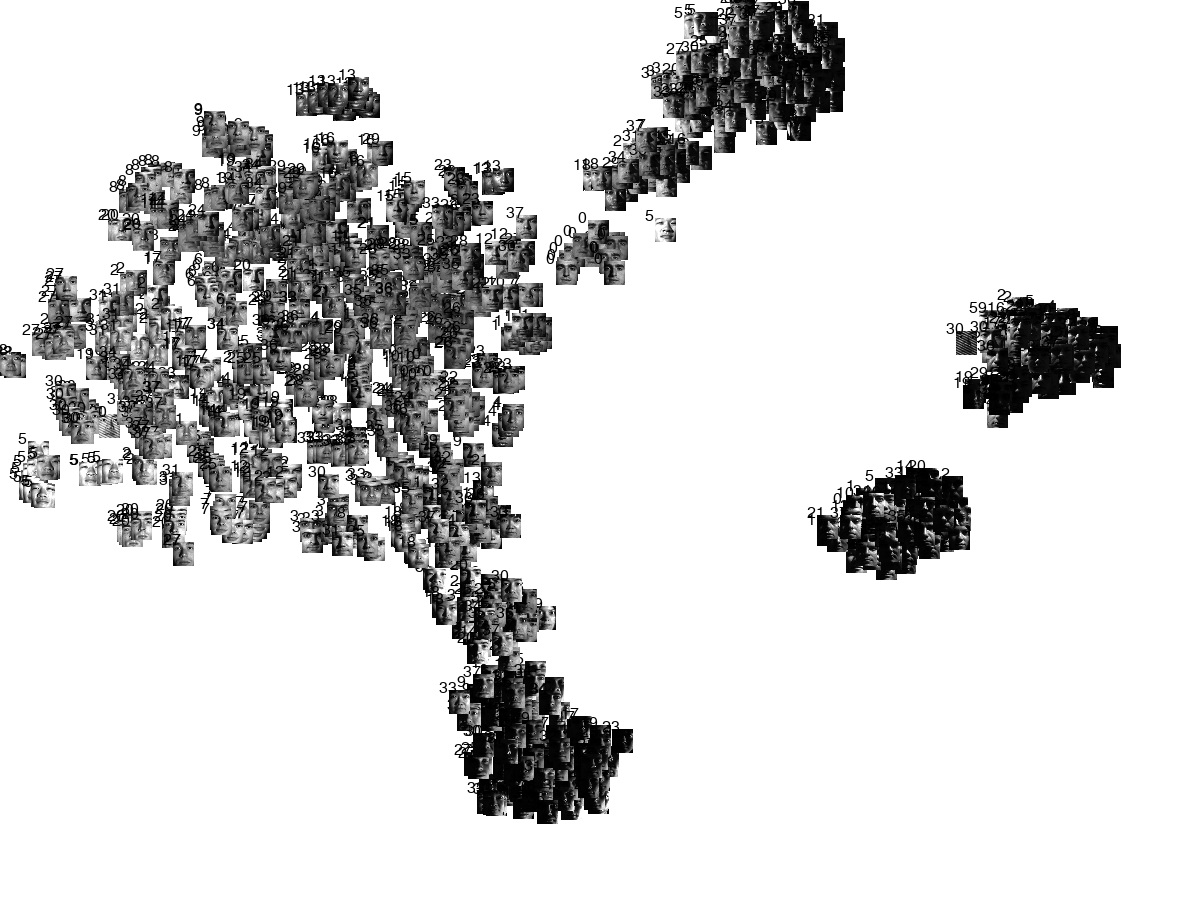}
        \caption{t-SNE on $\vect{z_S}$} \label{fig:compare1a}
    \end{subfigure}

    \vspace{0.2cm}
    \begin{subfigure}{0.31\textwidth}
        \includegraphics[width=1\textwidth]{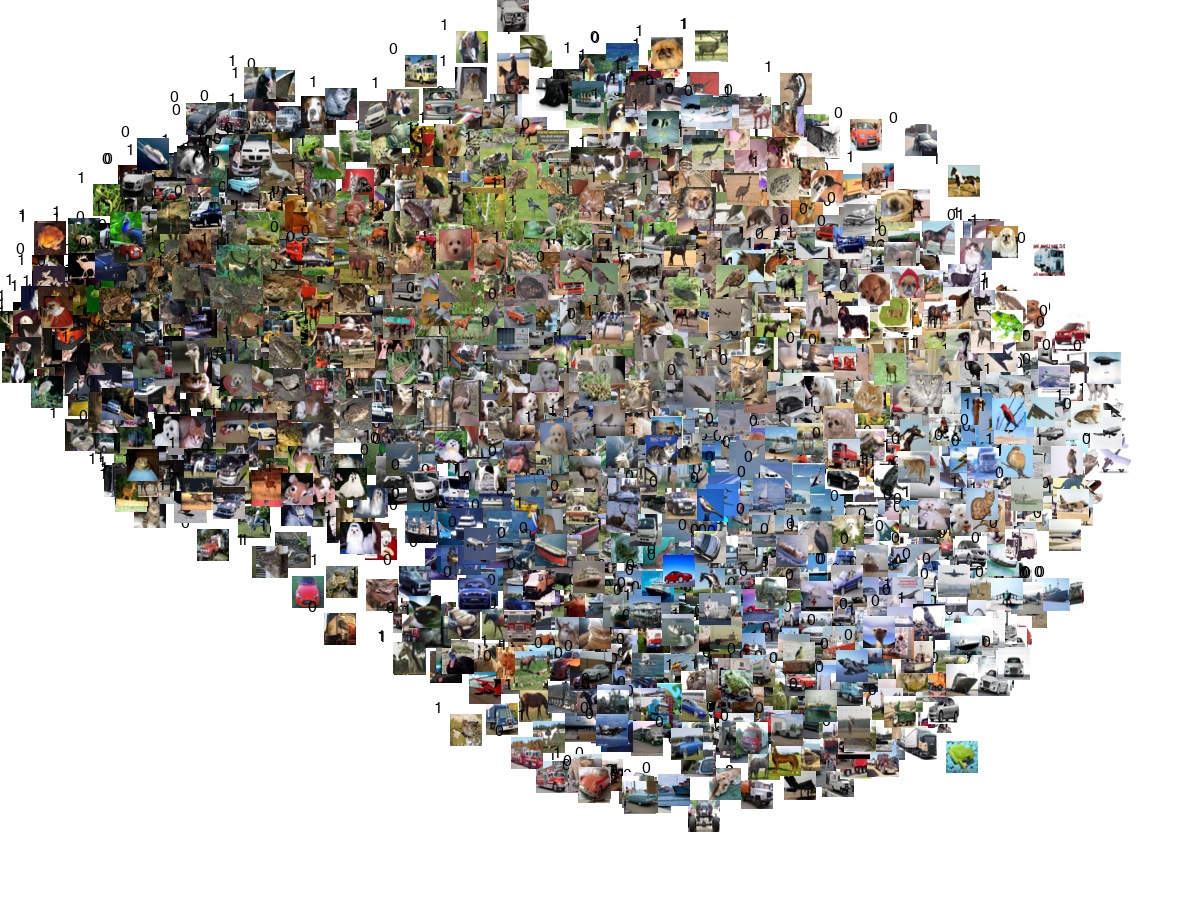}
        \caption{t-SNE on $x$} \label{fig:compare1a}
    \end{subfigure}
    \hfill
    \begin{subfigure}{0.31\textwidth}
        \includegraphics[width=1\textwidth]{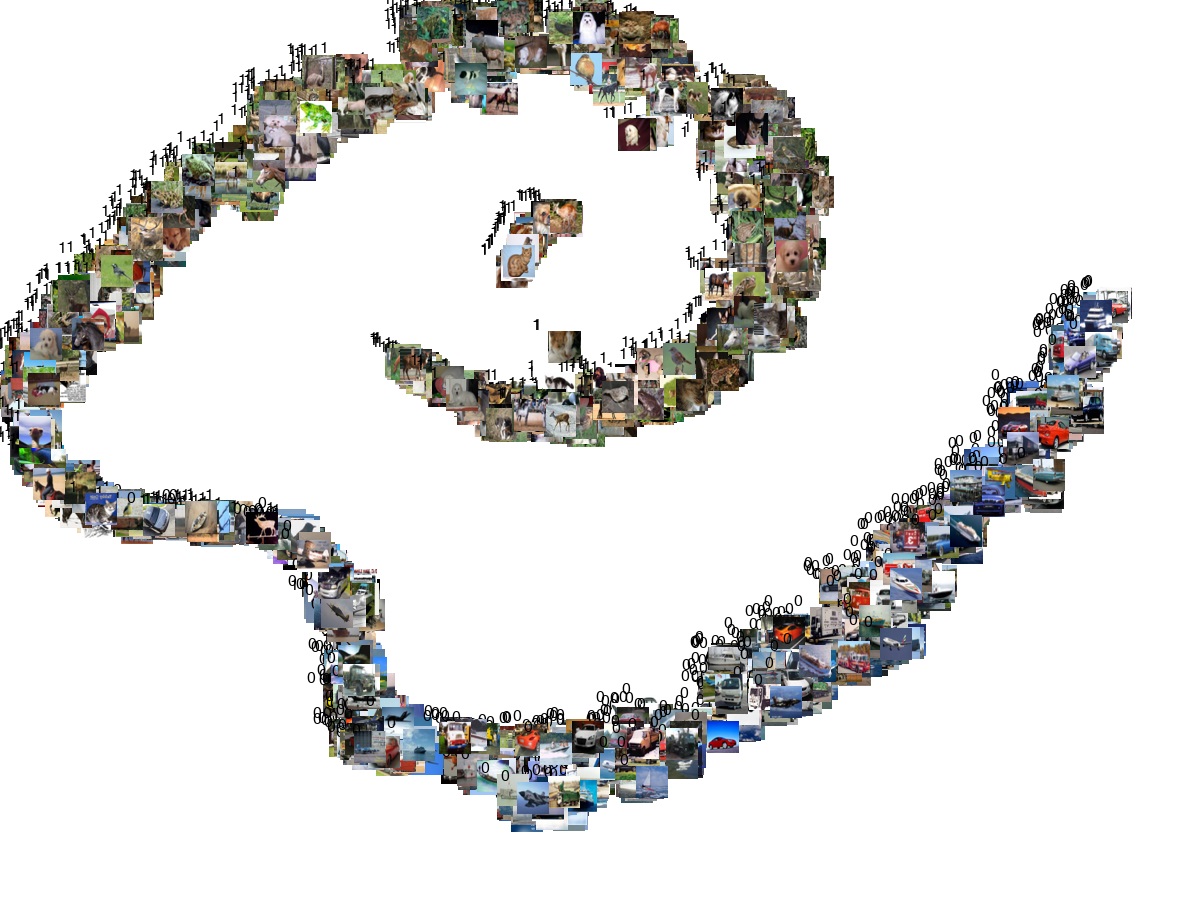}
        \caption{t-SNE on $\vect{z_T}$} \label{fig:compare1a}
    \end{subfigure}
    \hfill
    \begin{subfigure}{0.31\textwidth}
        \includegraphics[width=1\textwidth]{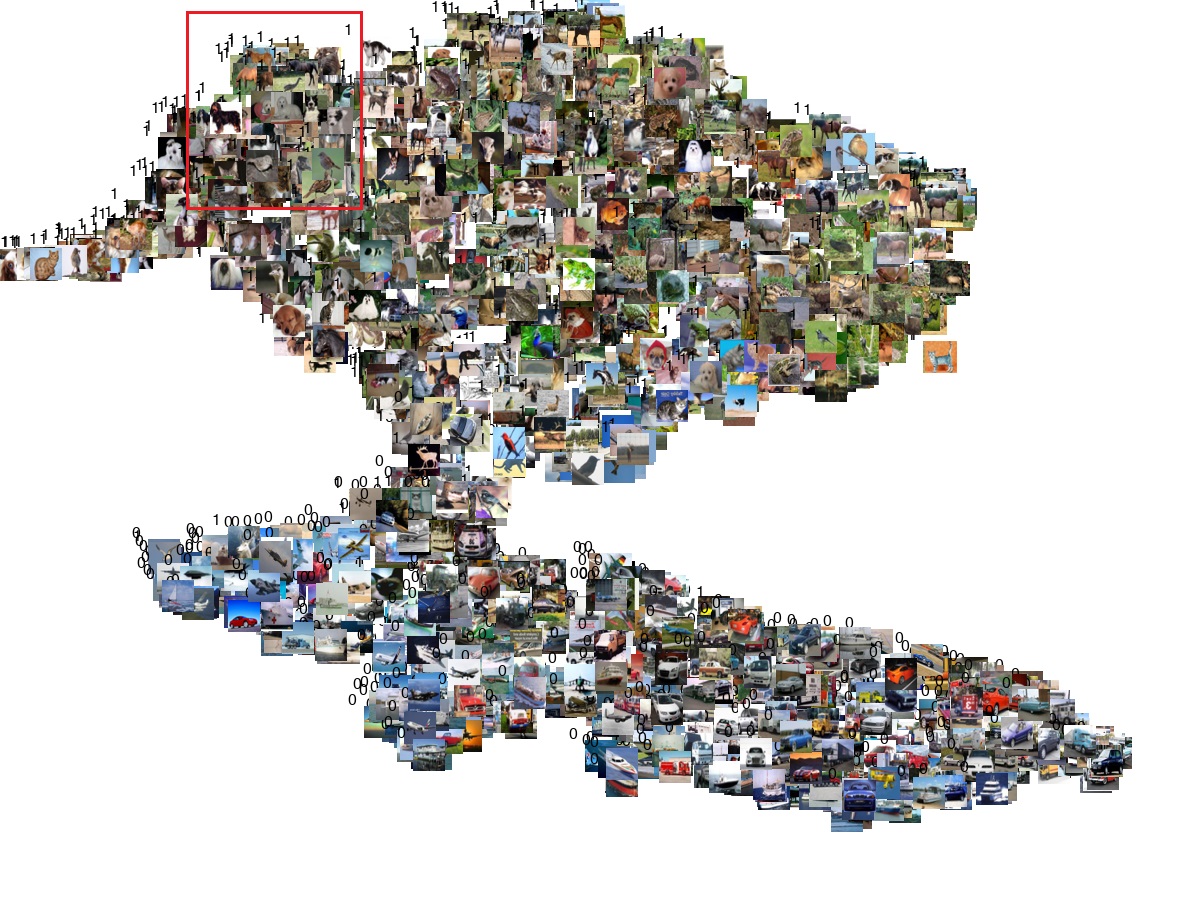}
        \caption{t-SNE on $\vect{z_S}$} \label{fig:compare1a}
    \end{subfigure}
    \caption{t-SNE visualization of the extended YaleB faces (top) and CIFAR-10 (bottom) images. Figure is better seen in color and high resolution.}    
    \label{fig:tsne}
\end{figure}

We visualize the learned embeddings using t-SNE  \cite{maaten2008visualizing} projections for the extended YaleB and CIFAR-10 datasets (\cf Fig.~\ref{fig:tsne}. We use the image space, $\vect{z_T}$, $\vect{z_S}$ as inputs to the projection to visualize what type of information is held within each representation. We also show the label of each image with regards to the target task to make it easier to investigate the clusters. For the extended YaleB, we see that, using the image space $\vect{x}$, the images are clustered mostly depending on their illumination conditions. However, when using $\vect{z_T}$, the images are not clustered according lighting conditions but rather, mostly based on the subject identity. Moreover, the visualization of representation $\vect{z_S}$ shows that the representation contains information about the sensitive class. For the CIFAR-10 dataset, using the image space basically clusters the images on the dominant color. When using $\vect{z_T}$, it is clear that the target information is separated where the right side represent the non-living objects, and the left to inside part represents the living objects. What should be observed in $\vect{z_T}$, is that within each target class, the fine classes are mixed and indistinguishable as we see cars, boats and trucks mixed in the right hand side of the figure, for example. The representation $\vect{z_S}$ has some information about the target class and also has the residual information about the fine classes as we see in the annotated red rectangle. A group of horses images are clustered together, then few dogs' images are clustered under it, then followed by birds. This shows that $\vect{z_S}$ has captured some sensitive information while $\vect{z_T}$ is more agnostic to the sensitive fine classes.

\subsection{Sensitivity analysis}
\label{sec:sensitivity}
\begin{figure}[t]
    \centering
    \includegraphics[width=0.23\textwidth]{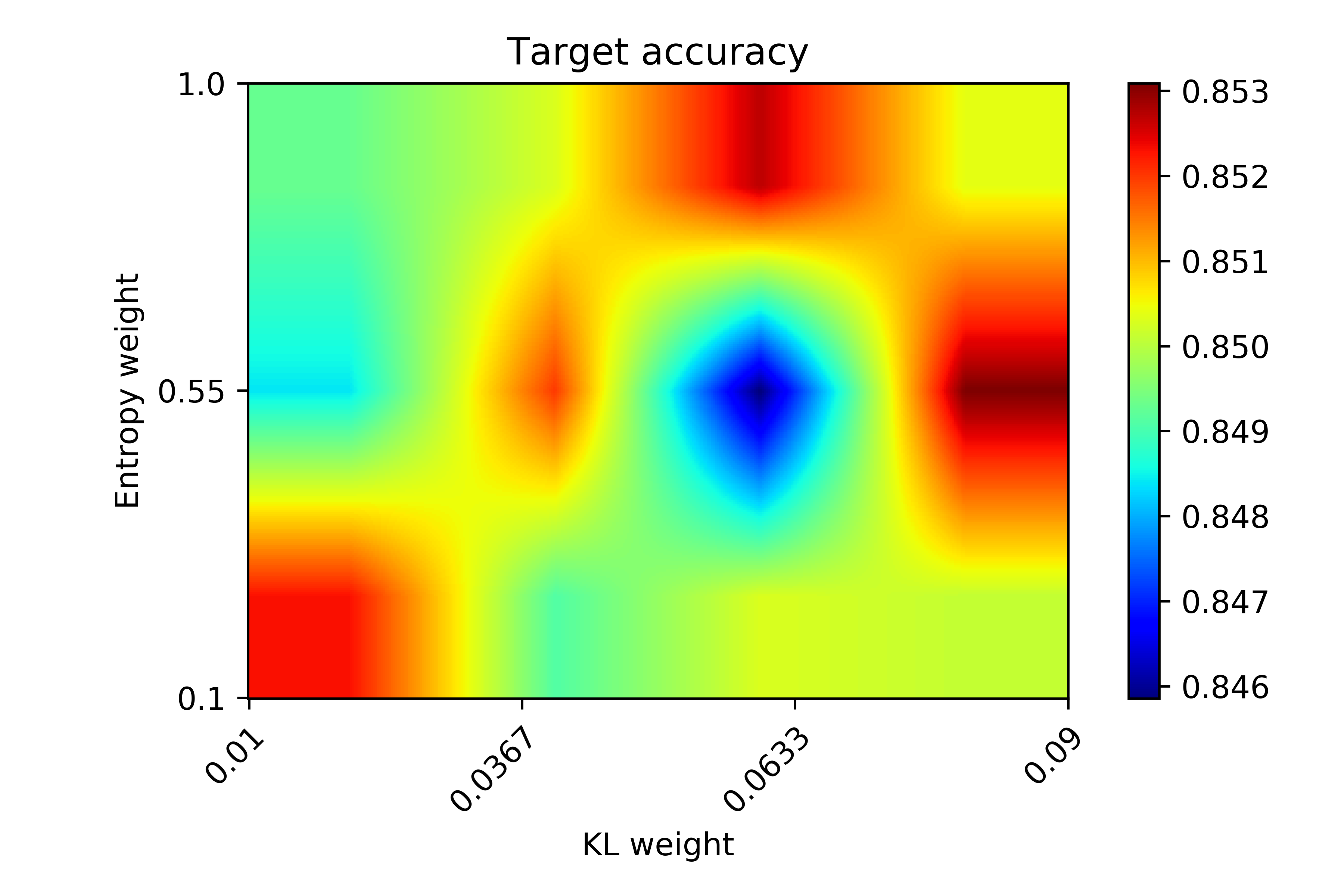}\hfill
    \includegraphics[width=0.23\textwidth]{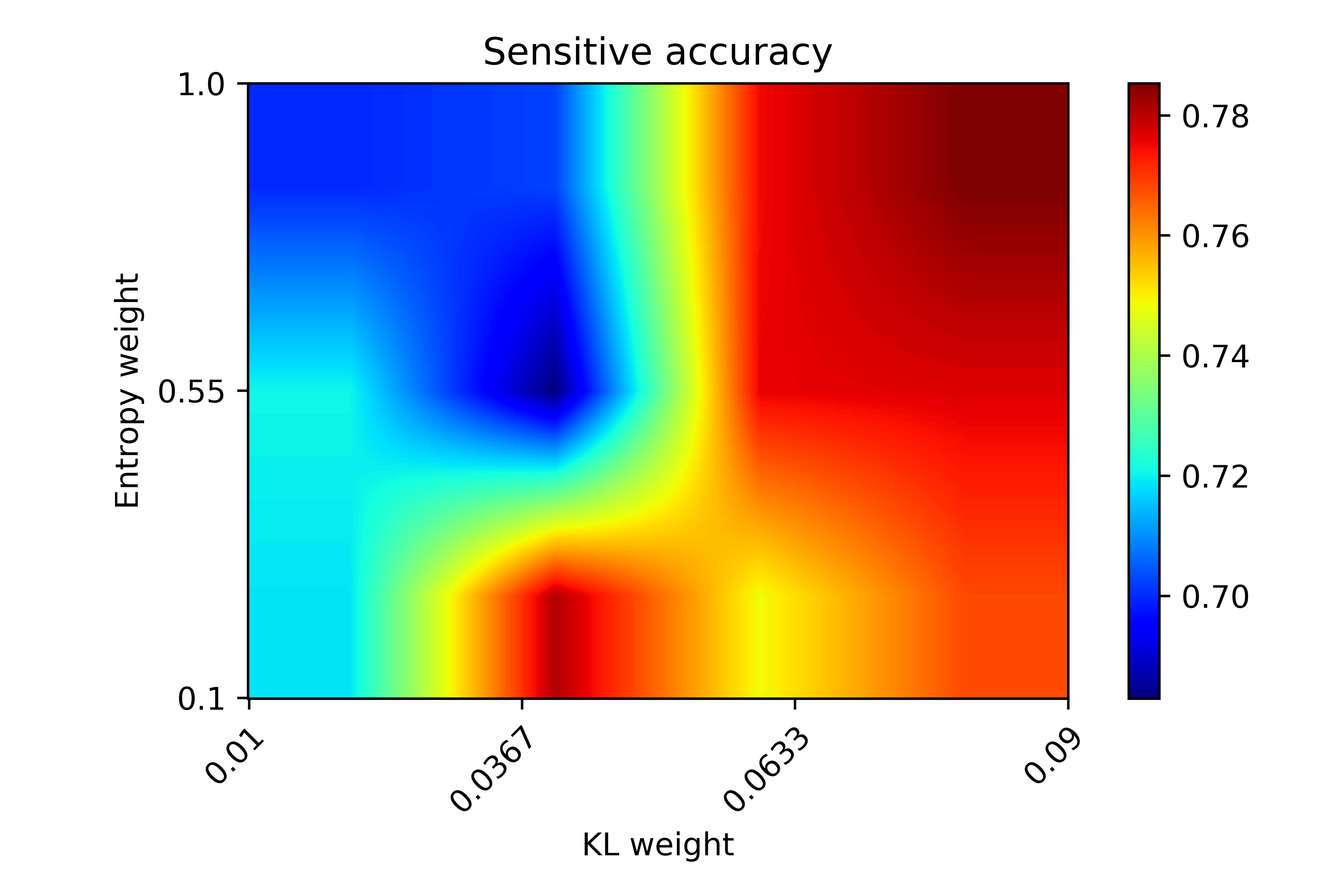}\hfill
    \includegraphics[width=0.23\textwidth]{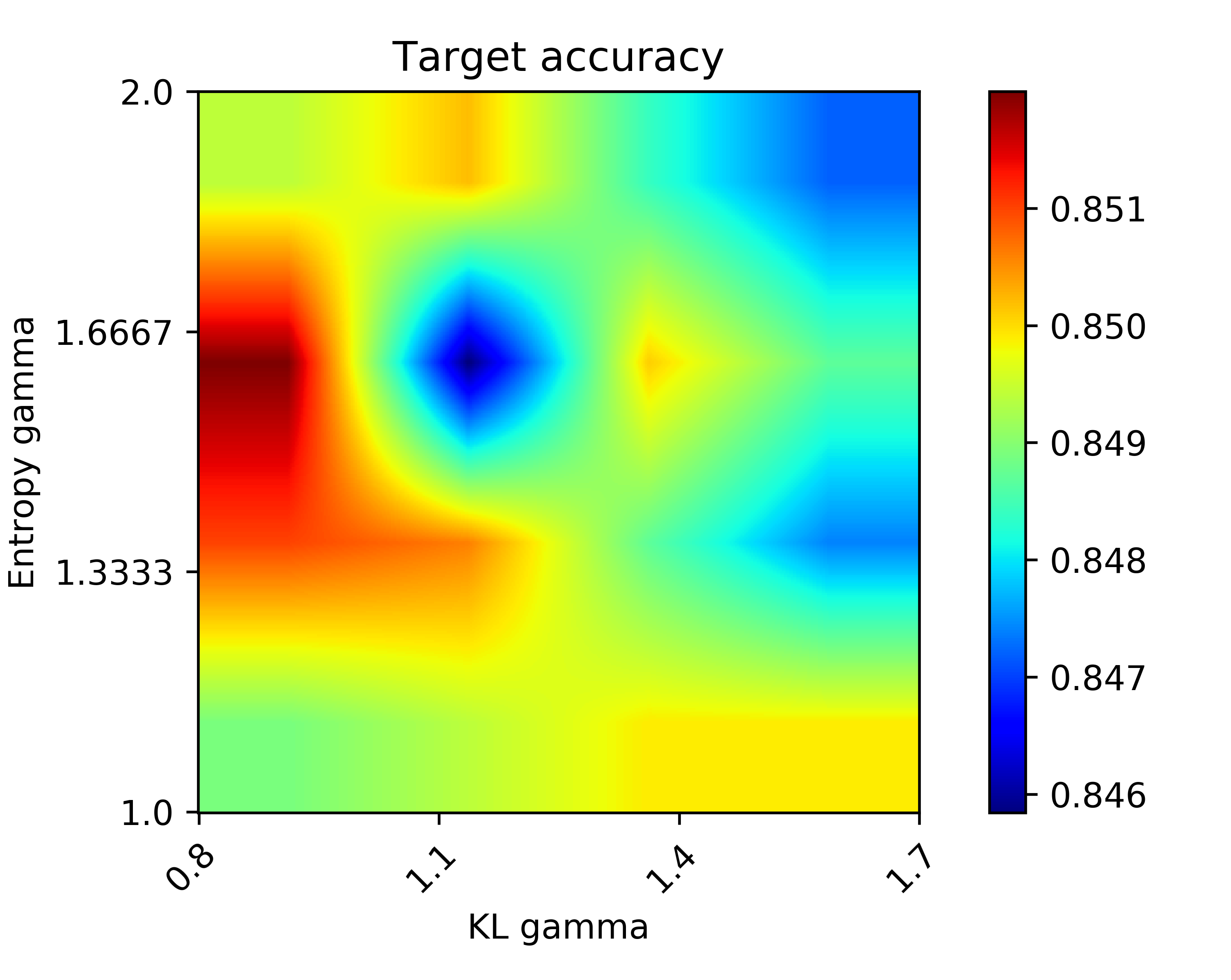}\hfill
    \includegraphics[width=0.23\textwidth]{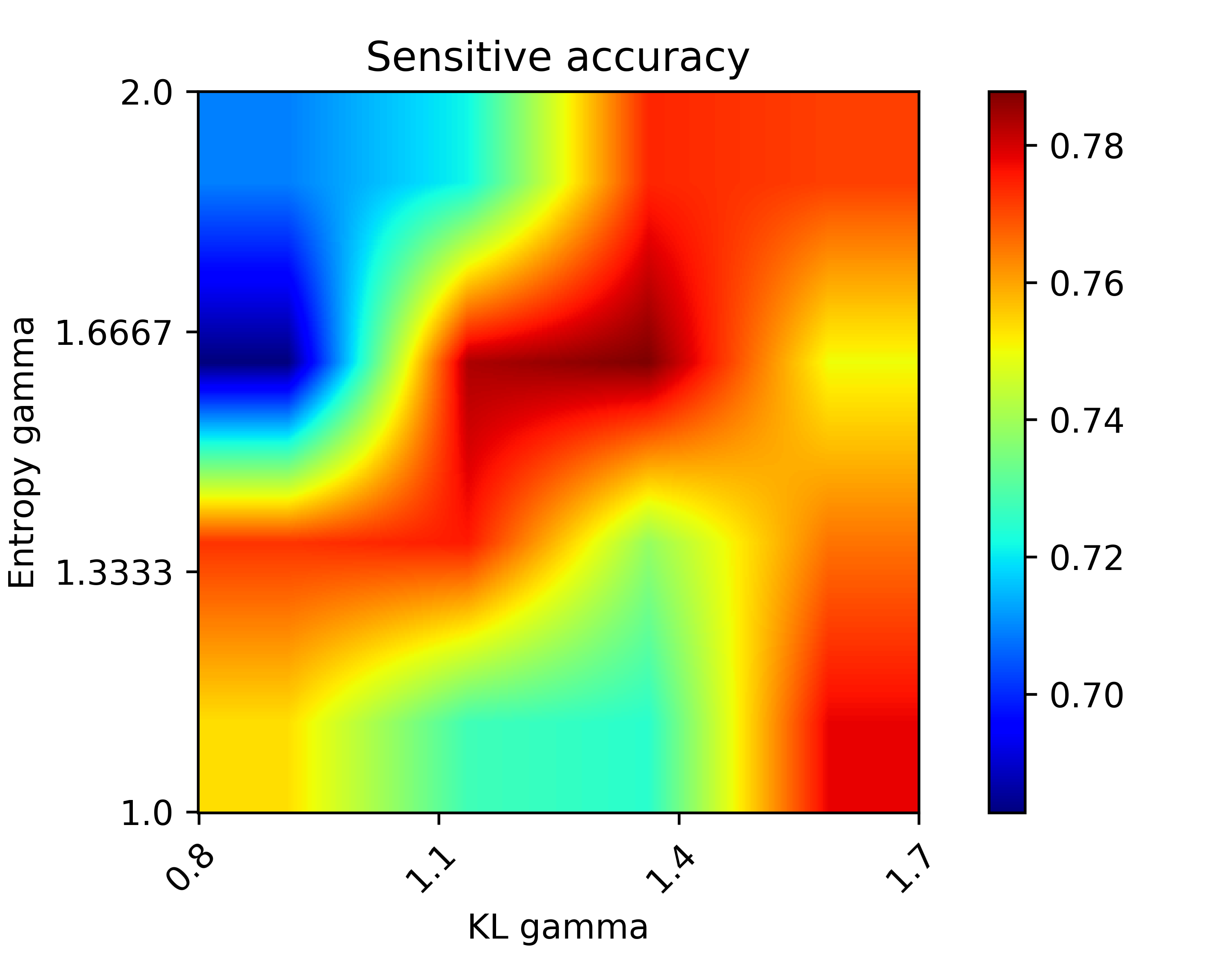}\hfill
    \caption{Sensitivity analysis on the Adult dataset}
    \label{fig:german-sensitivity}
\end{figure}
To analyze the effect of hyper-parameters choices on the sensitive and target accuracy, we show heatmaps of how the performance changes when the studied hyper-parameters are changed. The investigated hyper-parameters are KL weight ($\lambda_{OD}$), Entropy Weight ($\lambda_E$), KL gamma ($\gamma_{OD}$), and Entropy gamma ($\gamma_E$). We show the results on the Adult dataset. We can see that the sensitive accuracy is sensitive to $\lambda_{OD}$ more than $\lambda_{E}$ as changes in $\lambda_{E}$ do not induce much change on the sensitive accuracy. A similar trend is not visible on the target accuracy. Regarding the choice of $\gamma_{OD}$ and $\gamma_{E}$, we can see that the sensitive leakage is highly affected by these hyper-parameters and the results vary when changed. However, a more robust performance is observed on the target classification task.

\section{Conclusion}
In this work, we have proposed a novel model for learning invariant representations by decomposing the learned codes into sensitive and target representation. We imposed orthogonality and disentanglement constrains on the representations and forced the target representation to be uninformative of the sensitive information by maximizing sensitive entropy. The proposed approach is evaluated on five datasets and compared with state of the art models. The results show that our proposed model performs better than state of the art models on three datasets and performed comparably on the other two. We observe better hiding of sensitive information while affecting the target accuracy minimally. This goes inline with our hypothesis that decomposing the two representation and enforcing orthogonality could help with problem of information leakage by redirecting the information into the sensitive representation. One current limitation of this work is that it requires a target task to learn the disentanglement which could be avoided by learning the reconstruction as an auxiliary task. 

\section*{Acknowledgments}
S.A. is supported by the PRIME programme of the German Academic Exchange Service (DAAD) with funds from the German Federal Ministry of Education and Research (BMBF).

%
%
\bibliographystyle{splncs04}
\bibliography{egbib}
\end{document}